# Designing Domain-Specific Large Language Models - The Critical Role of Fine-Tuning in Public Opinion Simulation

**Haocheng Lin**

**Department of Computer Science, University College London, UK**

**Abstract**

Large language models (LLMs) have transformed natural language processing, yet face challenges in specialized tasks such as simulating opinions on environmental policies. This paper introduces a novel fine-tuning approach that integrates socio-demographic data from the UK Household Longitudinal Study, uniquely using profiling factors, such as age, gender, income, education, and region. This method enhances the accuracy and representation of generated views. By emulating diverse synthetic profiles, the fine-tuned models significantly outperform pre-trained counterparts, achieving measurable improvements in capturing demographic nuances. Evaluation metrics, including Chi-Squared, Cosine Similarity, Jaccard Index, and KL-divergence, reveal a strong alignment between synthetic and real-world opinions. This work demonstrates the potential of fine-tuned LLMs tailored to societal contexts to enable more ethical and precise policy simulations. Its broader implications include deploying LLMs in domains like healthcare and education, fostering inclusive and data-driven decision-making in both research and practice.

# 1 Introduction

## 1.1 Overview of Large Language Models and Their Growing Application Across Industries

Large Language Models (LLMs), such as GPT-4 and BERT, have revolutionized natural language processing (NLP) by performing complex tasks, such as generating human-like text, translating languages, summarizing lengthy documents, and understanding context to engage in diverse conversations. These models are trained on vast amounts of data, including the UK Household Longitudinal Study (UKHLS). Billions of parameters support them to execute domain-specific tasks across various industries, including healthcare, finance, legal services, and education [1, 2]. For instance, GPT-4, a transformer-based multimodal model, has been applied in areas, such as image processing, dialogue systems, and machine translation [3]. Its predecessor, GPT-3, with 175 billion parameters, serves as a comparative baseline to evaluate how GPT-4 evolved its capability to manage generalized tasks with minimal supervision [4]. Despite their versatility, general-purpose LLMs often struggle in specialized domains, such as public policy or environmental governance, which require deeper interpretation. These models are limited in predicting fringe socio-political opinions and need more context for integrating socio-demographic and policy-specific factors into creating more realistic synthetic profiles [5].

## 1.2 Limitations of General-Purpose LLMs in Domain-Specific Tasks

While pre-training on extensive public-domain datasets ensures that the LLMs could successfully perform general purpose tasks, this approach prevents them from excelling in specialized tasks, such as analysing diverse opinions on environmental policies [6]. Pre-trained models often fail to capture how socio-demographic factors influence individual responses to environmental issues.

For instance, previous research show that LLMs often overestimate the proportion of respondents concerned about climate change, resulting in misaligned policies that deepen socio-political divisions [7, 8]. Fine-tuning models with domain-specific datasets enables them to interpret the relationships between socio-demographic variables and expected responses, which is essential for accurately representing underrepresented groups.

Pre-trained models can amplify biases present in their training data, leading to skewed or inaccurate predictions when simulating opinions. For example, the COMPAS decision-making algorithm was criticized for reinforcing historical biases by disproportionately labelling individuals from marginalized groups as high-risk [9]. A similar risk exists with LLMs, which can perpetuate biases if the training data overrepresents certain narratives. While COMPAS makes categorical decisions, LLMs generate text that may reflect these biases unless specialized fine-tuning and prompt engineering are applied. To mitigate such biases, techniques like adversarial de-biasing and fairness constraints are critical to ensuring accurate, representative outputs in real-world applications [10].

## 1.3 Fine-Tuning as a Solution for Domain-Specific Adaptation

Fine-tuning offers a robust method to overcome the limitations of general-purpose LLMs by adapting pretrained models to domain-specific contexts. For instance, the UKHLS dataset, which provides comprehensive socio-demographic coverage, allows models to generate responses reflecting contemporary attitudes on real world issues. Through an iterative trial-and-improvement approach focusing on profiling variables such as age, income, education, and region, fine-tuning addresses the lack of contextually sensitive outputs in pre-trained models [11].

By enabling LLMs to account for socio-demographic factors, fine-tuning ensures the generation of responses that represent public sentiment across diverse groups, rather than reflecting only dominant opinions. This approach offers a cost-efficient method for gathering insights to improve environmental policies. Furthermore, fine-tuning enhances the scalability and applicability of LLMs, providing a more efficient solution than traditional models.

## 1.4 Objectives

This paper explores how fine-tuning enhances LLMs' performance in domain-specific tasks, particularly in simulating opinions on environmental policies. Using the UKHLS dataset as the primary conditioning and fine-tuning resource, this study aims to improve the accuracy of LLMs in predicting opinions and benchmark their outputs against real-world data. The study achieves these goals through the following objectives:

- Enhance LLMs' prediction accuracy by fine-tuning them with the UK Household Longitudinal Study (UKHLS) datasets.
- Define benchmarks for evaluating fine-tuned models' response distributions against real-world public opinion data using metrics, such as Chi-Square test scores, Cosine Similarity, and Jaccard Index.
- Demonstrate fine-tuning's ability to represent diverse public opinions on environmental policies, focusing on complex socio-demographic variables.
- Identify the limitations of fine-tuning, including overfitting, bias, and computational costs, to inform policymakers and improve public engagement with environmental issues.

By addressing these objectives, this study provides policymakers and developers with a scalable and ethical approach to simulate public opinions, thereby supporting the development of inclusive, data-driven policies that align with societal expectations [12].

# 2 Literature Review

## 2.1 LLM Architecture: Development of Transformer-Based Models

Large Language Models (LLMs), such as GPT-4 and BERT, use a transformer architecture, which excels in processing large datasets [13]. Unlike traditional models like Recurrent Neural Networks (RNNs) or Long Short-Term Memory (LSTM) networks, which process input sequentially, transformers use a self-attention mechanism that performs in parallel. This design allows transformers to scale efficiently, capture long-range dependencies, and overcome the limitations of vanishing gradients that hinder the performance of sequential models in deep networks. Vanishing gradients occur when the gradients become extremely small during backpropagation, making it difficult for models to learn long-range dependencies. By contrast, transformers process all tokens in parallel, enabling them to maintain gradient stability and learn more effectively from long sequences.

In transformer-based models, their architecture consists of two main components, the self-attention layers and feed-forward networks. The self-attention mechanism enables the model to evaluate the significance of each token in relation to all other tokens within the sequence simultaneously. This capability allows transformers to effectively capture relationships and context, irrespective of whether the tokens are close or distant within the input sequence. Self-attention layers allow models to focus on relevant words across the entire text, enriching its ability to process and understand nuanced content. Both measures can add more contextual information to support NLP for processing and building relevant profiles to understand why people have different attitudes on environmental issues. The transformers excel in various NLP tasks, including language modeling, translation, and text summarization, making them indispensable in modern applications.

Each attention weights are calculated using the query, key, and value vectors for quantifying the focus in the input sequence. These variables quantify how the transformer's attention head focuses on each input sequence token. There are multiple attention heads in the process layers, with each head focusing on distinct aspects; for example, one head may prioritize syntactical relationships, while another focuses on semantic connections. Multi-head structure enables the model to process language from multiple abstraction levels. After some calculations, the weights are passed through a feedforward network, consisting of multiple transformer and attention layers. For example, the transformers use a multi-head attention mechanism, which enables the model to capture different type of relationships between profiling variables. Having these layered structures allow the models to understand different abstraction levels in parallel, which improve their ability at interpreting the overall context behind the expected output distributions.

Transformers can process and retain information from big complex datasets, contributing to the development of accurate LLMs, such as GPT-4, BERT, and T5, which makes them capable of performing a broad scope of language tasks with minimal supervision [2, 4]. BERT's bidirectional training considers context from both directions in the text, improving its perception in deep contextual cases. GPT uses probability sequences to predict the next word in the output, enhancing its ability to produce and simulate a broad scope of synthetic opinions on environmental issues. Both examples feature how transformer models are scalable and flexible enough to perform specialised and domain-specific tasks.

## 2.2 Fine-Tuning Techniques: Adapting LLMs for Domain-Specific Applications

### 2.2.1 Fine-tuning Overview

Fine-tuning involves adapting pre-trained Large Language Models to perform effectively in specialized domains by retraining them on a smaller, domain-specific dataset. This process aligns the model's outputs with the expected distributions of the target domain. A successful fine-tuning strategy incorporates adjustments to the model's parameters, including weights, learning rate, batch size, and the number of epochs, to ensure optimal performance [11].

#### 2.2.2 Efficient Fine-tuning with Adapters and LoRA

Recent advancements, such as Adapters and LoRA (Low-Rank Algorithm), allow fine-tuning to focus on a small subset of training data at a time, significantly improving the models' efficiency [14].

For example, adapters are light-weight modules inserted between transformer layers and retrained for specific tasks. This modular design allows models to adapt to task-specific requirements while retaining their general capabilities. In this study, adapters enable a more flexible adjustment to reflect the socio-demographic profiles, in particular the younger or lower-income groups.

LoRA introduces low-rank matrices into transformer layers, which reduces the number of training parameters. It is a cost-effective alternative for fine-tuning smaller and domain-specific subsets without compromising the performance, which is crucial for balancing general knowledge retention with task-specific accuracy.

Both techniques have their limitations: while adapters cause some latency, LoRA's low-rank approximations underperform in complex task scenarios. Complementing the methods by combining the modular structure of adapters with LoRA enables the designed LLMs to achieve a greater degree of flexibility.

#### 2.2.3 Task-specific Fine-tuning and Prompt Engineering

Task-specific fine-tuning modifies the LLMs' internal parameters to ensure their alignment with the specific task requirements [5]. In contrast, prompt engineering crafts the instructions to guide the LLMs without altering the structure. The designed prompts are classified into the following two categories [15]:

- **System prompts:** Defines the persona for the LLMs to understand the operational context.
- **User prompts:** Specifies the instructions for the LLMs to execute, which contains the relevant context and descriptions of the expected output.

This approach is effective in few- or zero-shot learning scenarios by enabling the models to generalise from a limited set of examples and adapt to new tasks without using an extensive amount of labelled data. By using carefully designed prompts, LLMs could ensure that their outputs are tailored for task-specific requirements, which enhances the performances in fields, such as sentiment analysis, classification, and decision-making.

However, prompt engineering and fine-tuning serve different purposes and can be combined to optimize the LLMs' performance. While fine-tuning adjusts the models' inner parameters based on a training dataset, prompt engineering guides the models into producing desired outputs by providing them with interpretable instructions. For example, the T5-base prompt-learning paradigm optimises prompts to ensure that the LLMs could achieve at least 75% accuracy level of a fully fine-tuned LLM model [16].

#### 2.2.4 Integrating Socio-Demographic Factors

The UK Household Longitudinal Study Datasets effectively conditions LLMs to understand public attitudes toward environmental policies, including climate change, renewable energy, and sustainability plans [17]. Including socio-demographic factors help LLMs to produce accurate, context-aware predictions. Profiling variables align model responses with real-world contexts and reveal factors that shape the public's perceptions of environmental policies. This knowledge can help policymakers with designing more inclusive and transparent ideas that not only solves environmental challenges but gains enough support to be passed through the legislations.

### 2.3 Ethical and Domain-Specific Considerations

#### 2.3.1 Addressing Biases in LLM Outputs

There are some ethical challenges when fine-tuning LLMs, such as emphasizing the risks of bias. Pre-trained LLMs, such as GPT-3 and BERT, inherit biases from their large-scale training datasets. Biases, such as those related to gender, race, or socio-economic status, could skew the LLMs' response distributions [18]. During pre-training, the LLM-generated answers overrepresented support for renewable energy from the rural demographics. A detailed demographic parity test identified the source of the biases, enabling oversampling to correct the representation of different stakeholder groups. Preprocessing techniques, such as normalization and sampling, help to mitigate the risks of bias (see Section 2.3.4). For example, the UK Household Longitudinal Study (UKHLS) has included booster samples of 6500 participants for ensuring an accurate representation on a national scale. New households were sampled to replace the inactive households who either moved out or stopped participating in the study.

If the fine-tuning data is not representative enough, it could reinforce existing biases [10]. This amplification can have serious real-world consequences, especially in domains, such as public policy, where AI-driven simulations can influence decision-making. Unlike traditional methods, adversarial debiasing is integrated into fine-tuning for identifying potential bias patterns, ensuring the LLM-generated responses are inclusive and equitable. Bias identification uses a secondary model to detect and address the sensitive attributes that influences predictions. The learning process helps the LLMs reduce vulnerabilities associated with sensitive variables. However, adversarial debiasing can suppress regional variations, which requires future studies to explore adaptive debiasing to balance fairness with representing demographic differences.

### 2.3.2 Impact of Echo Chambers

In public opinion simulations, it is crucial to ensure that fine-tuning captures a comprehensive spectrum of public sentiment across diverse demographic groups. Research indicates that lower socio-economic groups are underrepresented in opinion-based research studies [19]. Factors, such as digital divides, time constraints, mistrust of research, and cultural barriers, contribute to their limited participation. These gaps pose a significant risk, as they might lead to policies that disproportionately benefit the wealthier demographics, reinforcing existing inequalities.

Echo chambers distort the simulation data, leading to designed models that overrepresent extreme views. Social media exacerbates this echo chamber effect, which promotes contents that align with the users' preferences and reinforcing their pre-existing beliefs. For example, a study of climate change discussions in Ireland found that forums used to study policy attitudes were biased due to the polarized nature of conversations [20]. Similarly, a UK case study on news consumption revealed three distinct groups: 22% were 'news lovers,' consuming content from multiple sources, 55% were 'daily briefers,' relying on fewer sources, and 23% were 'casual users,' with minimal exposure to daily news. These patterns suggest that while most users are exposed to a variety of viewpoints, a sizeable portion of the population remains confined to narrow and homogeneous media. These biases risk generating policies that fail to address the population's needs.

### 2.3.3 Mitigating Polarization with Diversity-Enhancing Algorithms

When conducting surveys on political preferences, only 2% to 5% of online users identify themselves as members of left- or right-wing echo chambers. However, these responses often reflect biases introduced by users who have been desensitized through repeated exposure to ideologically aligned content. For instance, the echo chamber effect explains why individuals with pro-environmental attitudes are more likely to participate in studies on environmental issues, resulting in overrepresentation biases from generic pre-trained LLMs that favour pro-environmental policies. Diversity-enhancing algorithms, such as cluster-based sampling, mitigate these biases by encouraging LLMs to generate outputs that reflect public opinion. Specifically, cluster-based sampling mitigates the underrepresentation of rural populations from the UKHLS dataset with supplementary samples. By correcting the imbalances, diversity-enhancing algorithms ensure that policy simulations are both inclusive and actionable, supporting equitable decision-making across diverse communities.

#### 2.3.4 Preprocessing Framework for Data Processing

To address potential biases, a set of careful data preprocessing is required to correct potential biases during fine-tuning [21]. To address these biases, preprocessing uses the following steps (Fig. 1):

1. Data cleaning removes irrelevant content, such as invalid values, linguistic errors, and random characters. For instance, missing UKHLS dataset values were imputed to ensure completeness, while random or corrupted entries were discarded.
2. Normalization standardizes text format for a better comparison between synthetic and real-world profiles. For example, the education categories simplify into "no formal education, secondary education, post-secondary education, and higher education".
3. Identifying and removing duplicate profiles to ensure an integral dataset without redundancies.
4. Balancing datasets by ensuring an equal representation of demographic groups (See Section 3.2).
5. Search for additional data to address underrepresentation from the minority groups.
6. Randomize data during each training cycle to prevent the model from memorising the data patterns.

After preprocessing, the percentage of invalid values in the UKHLS profiling dataset reduced from 27.5% to 0%, ensuring a more balanced representation of socio-demographic groups. For instance, normalization simplified profiling variables into narrower categories, which enables an easier visualisation and prevents the anomalies from distorting the profiles.

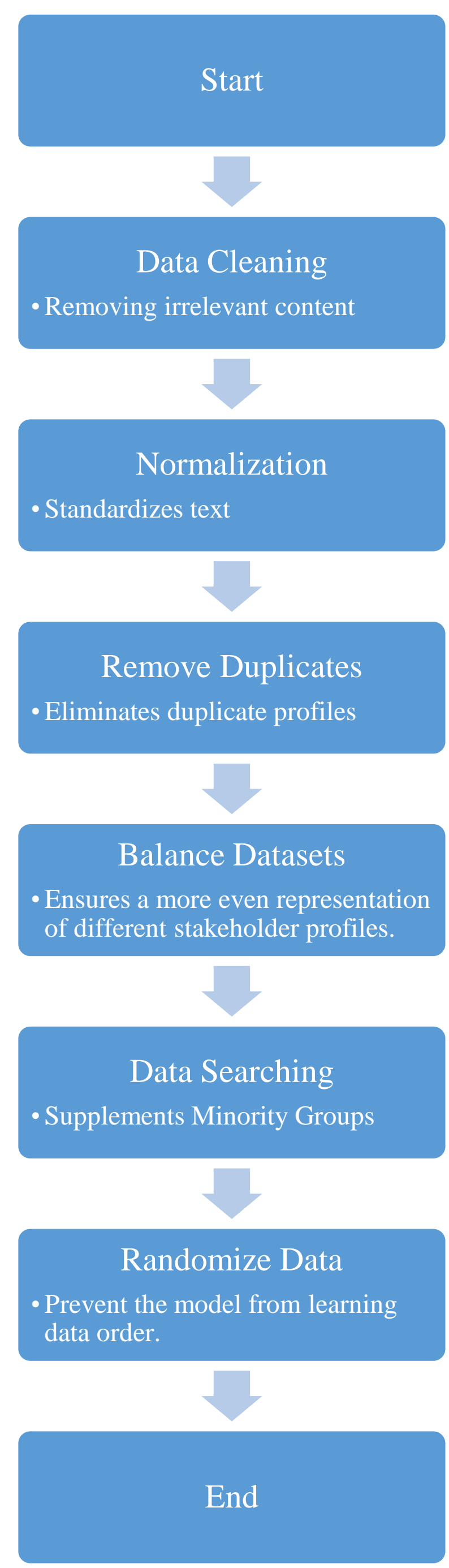

**Fig. 1** A flowchart visualising the preprocessing steps for mitigating any future biases in fine-tuning.

# 3 Methodology: Fine-Tuning LLMs for Public Opinion Simulation

## 3.1 Pre-Trained Model Overview

This study uses GPT-4 variants (GPT-4o, GPT-4o mini, and GPT-4o1-preview) to examine how fine-tuned models are specialised for simulating opinions on environmental policies. Prior to fine-tuning, these models can perform different natural language processing tasks due to their pre-training on large datasets. For example, GPT-3.5-turbo, with 175 billion parameters, is trained on diverse datasets, including open-source collections and web data [13, 2]. However, pre-trained models often struggle with tasks requiring domain-specific knowledge to reflect the subtle differences between opinions accurately. For fine-tuning, a grid search determines optimal hyperparameters, balancing between training time and model accuracy. An AdamW optimizer was used with a learning rate of $2\times10{-}5$ for allowing the model to converge efficiently while minimizing the risks of overfitting. A batch size of 32 balances computation efficiency with model performance. To further reduce the risks of overfitting, early stopping was set to 10 epochs, which halts training if there were no significant improvements and that the model could retain its generalizability.

While hyperparameter optimisation ensures that fine-tuned models achieve peak performance, it is equally important to validate their capabilities using real-world benchmarks to assess how suitable they are for domain-specific tasks. A comparative analysis of GPT-3.5 and GPT-4 performances using an interdisciplinary set of exam benchmarks highlights a substantial amount of improvements made by the newer GPT-4 model. These exam benchmarks encompass a broad scope of subjects, including maths, physics, history, and psychology. In every category, GPT-4 consistently outperformed GPT-3.5, achieving scores that reflect its ability at handling complex tasks. For instance, in a bar exam, GPT-4 ranked within the top 10% of the test participants, while GPT-3.5's performance languishes in the bottom 10% [3]. Similarly, GPT-4 could apply problem solving in domains, such as calculus, economics, and world history, adapting its ability to break down complex context and deriving unique solutions. These findings support this study's objectives 1 and 3 by confirming GPT-4's suitability at modelling socio-political and demographic factors in opinion simulation tasks [4].

## 3.2 Datasets: UK Household Longitudinal Study (UKHLS)

The UK Household Longitudinal Study Dataset fine-tunes selected LLMs, which contains an extensive panel survey tracking 40,000 households between January 2009 and May 2023 [22]. These profiling variables, such as age, gender, income, education, region, and family, offers insights into how the demographic structure is shifting and influencing attitudes to climate change and environmental policies over time [23]. SHAP (Shapley additive explanations) helps to select relevant profiling variables with the top N most relevant variables ranked by importance. Some of the environmental variables used in this study include:

- **Current lifestyle (scenv_crlf):** Determines if an individual's lifestyle is environmentally friendly and if they need to adopt pro-environmental behaviours.
- **Support for climate policies (scenv_pmep, orga3):** Quantifies the public support for environmental policies by gathering personal habit variables, such as willingness to pay for green tariffs and whether an individual is happy to devote time for volunteering in environmental groups.
- **Perceptions about climate change (scenv_meds):** Determines if an individual has a positive perception about the world's future in tackling environmental issues, influencing their mindset about whether they think it is worth the effort to implement green policies.

Limitations: The dataset's temporal nature, spanning over 2 decades, might not fully capture the subtle shifts during recent periods as technology advances at an exponential rate. Although imputation and oversampling aim to address missing data and sampling biases, these adjustments fail to account for non-response biases and require careful validation to ensure that the updated sample represents the population. The temporal nature of the UKHLS dataset necessitates regular updates to reflect evolving societal trends, ensuring the relevance of the fine-tuning models.

These environmental variables provide a multidimensional perspective on public opinions, making them ideal for fine-tuning LLMs into producing representative responses [17]. Table 1 highlights three sample synthetic profiles constructed from probability distributions of selected profiling variables, ensuring an accurate representation of the UK demographic structure. These profiles define the LLMs' role through system prompts, enabling the models to simulate nuanced opinions reflective of diverse socio-demographic contexts. Beyond environmental applications, the UKHLS dataset supports public health modelling (e.g., vaccine uptake by age groups) and economic planning from income distributions, establishing the LLMs as a foundational interdisciplinary knowledge base.

To prepare UKHLS data for fine-tuning, several preprocessing steps aim to ensure that the data is consistent and relevant. Firstly, imputation corrects the invalid profiling values to rebalance the demographic structure. Secondly, the variables' format is standardized to optimise the number of prompting tokens and to ensure the output distributions are comparable with each other. Lastly, sampling techniques, like the Synthetic Minority Over-sampling Technique (SMOTE) increase the proportion of underrepresented groups to ensure balance across socio-demographic groups. SMOTE generates synthetic minority samples by interpolating existing data points. Unlike the existing duplication techniques, SMOTE maintains original data distributions without introducing any redundant data. For instance, SMOTE could increase the proportion of the rural population, who might have less data than their urban counterparts.

**Table 1** Three examples of synthetic profiles for conditioning the LLMs into generating representative views.

| Profiles | Details | |
|---|---|---|
| **Profile 1** | Age Group | In terms of my age, my age group is 40 - 49. |
| | Gender | I am Male. |
| | Ethnic Group | Racially, I am British. |
| | Marital Status | My marital status is Married. |
| | Profession | My profession is Semi-Routine Occupations. |
| | Highest Qualifications | In terms of my qualifications, I do not have any qualifications. |
| | Monthly Income (£) | Financially, my monthly income is £3213. |
| | Living Area | I live in an urban area. |
| | Region | I live in the South East Region. |
| | Number of Children | I have 1 child. |
| | Voting Intention | Ideologically, I describe myself as a Green Party supporter. |
| **Profile 2** | Age Group | In terms of my age, my age group is 40 - 49. |
| | Gender | I am Female. |
| | Ethnic Group | Racially, I am British. |
| | Marital Status | My marital status is Divorced. |
| | Profession | My profession is Higher Professional. |
| | Highest Qualifications | In terms of my qualifications, My highest qualification is foundation. |
| | Monthly Income (£) | Financially, my monthly income is £9007. |
| | Living Area | I live in a rural area. |
| | Region | I live in the West Midlands Region. |
| | Number of Children | I do not have any children. |
| | Voting Intention | Ideologically, I describe myself as a Labour Party Supporter. |
| **Profile 3** | Age Group | In terms of my age, my age group is 40 - 49. |
| | Gender | I am Male. |
| | Ethnic Group | Racially, I am Indian. |
| | Marital Status | My marital status is Married. |
| | Profession | My profession is Lower Supervisory and Technical. |
| | Highest Qualifications | In terms of my qualifications, My highest qualification is university. |
| | Monthly Income (£) | Financially, my monthly income is £3795. |
| | Living Area | I live in an urban area. |
| | Region | I live in London. |
| | Number of Children | I have 2 children. |
| | Voting Intention | Ideologically, I describe myself as a Conservative Party Supporter. |

## 3.3 Fine-Tuning Process

Fine-tuning involves adapting the pre-trained LLMs to improve the simulation of opinions on environmental issues using the UKHLS dataset. During fine-tuning, the model parameters are weighted by the importance of selected features. Early stopping aims to mitigate overfitting, and dropout layers were added to reduce reliance on specific features. A regularisation term (0.01 weight decay) was applied to regularise the model throughout its training. Using transfer learning principles, fine-tuning helps the model to retain general knowledge while ensuring that it can specialise in understanding socio-demographic differences.

### 3.3.1 Multi-Objective Optimisation Framework

In a multi-objective optimisation framework, the loss function is updated as weighted combination of three components, performance, fairness, and bias reduction losses (Eq. 1).

$$L = \alpha \cdot L_{\text{performance}} + \beta \cdot L_{\text{fairness}} + \gamma \cdot L_{\text{bias reduction}} \quad (1)$$

Performance loss ensures that the model remains accurate during task-specific situations by measuring the differences between predicted and expected distributions. Traditional supervised learning approaches help fine-tuning with capturing accurate responses. Examples of loss functions, including the cross-entropy and mean-squared losses:

- Cross-entropy loss measures the difference between expected and synthetic responses for categorical variables.
- Mean-squared error quantifies average differences on a continuous scale, minimizing the losses to ensure that the designed models can emulate the expected response distributions.

These components maintain the models' ability at generalizing to socio-demographic differences between each synthetic profile, which ensures that the fine-tuned models could become more successful in the domain-specific areas. Fairness loss quantifies differences between model outputs by socio-demographic group. These examples of quantifiers include demographic parity difference (DPD) and equal opportunity difference (EOD). DPD compares positive outcome rates to quantify difference across groups; while EOD determines if each stakeholder group have an equal amount of opportunities for achieving positive outcomes. Bias reduction loss mitigates biases by performing adversarial debiasing for improving the representation of underrepresented groups.

**Table 2** Empirical results of 10 samples, $\alpha$, $\beta$, and $\gamma$ hyperparameters, for the selected question "Please select the extent to which you agree or disagree with the following statement: My behaviour and everyday lifestyle contribute to climate change." Each result derives from a grid search method, which evaluates multiple combinations of $\alpha$, $\beta$, and $\gamma$ to determine the optimal parameters.

| Sample | Alpha (Performance) | Beta (Fairness) | Gamma (Bias Reduction) |
|---|---|---|---|
| Sample 1 | 0.41 | 0.2 | 0.22 |
| Sample 2 | 0.59 | 0.39 | 0.13 |
| Sample 3 | 0.52 | 0.37 | 0.16 |
| Sample 4 | 0.48 | 0.24 | 0.17 |
| Sample 5 | 0.35 | 0.24 | 0.19 |
| Sample 6 | 0.35 | 0.24 | 0.26 |
| Sample 7 | 0.32 | 0.26 | 0.14 |
| Sample 8 | 0.56 | 0.3 | 0.2 |
| Sample 9 | 0.48 | 0.29 | 0.22 |
| Sample 10 | 0.51 | 0.26 | 0.11 |

Its component measures the probability that an attribute can be selected as a profiling variable as a part of the synthetic profile. Alongside heatmaps, these probabilities help to select more relevant and less sensitive profiling variables, which improves the models' accuracy, equity, and inclusivity.

From Eq. 1, $\alpha$, $\beta$, and $\gamma$ measure the degree to which hyperparameters need to be tuned for balancing trade-offs. Each component is aggregated in the developed loss function for a weighted dynamic calculation, which maintains performance without compromising fairness and bias.

### 3.3.2 Adversarial debiasing

Adversarial debiasing trains using sensitive attributes, such as age or income, to discourage the models from generating discriminative outcomes against these protected features. An example loss function could be expressed as the negative of a logarithm of the predicted probability of the sensitive attribute (Eq. 2).

$$L_{\text{Bias Reduction}} = -\log(p(\hat{y}_{\text{sensitive}})) \quad (2)$$

### 3.3.3 Illustration of Fine-tuning Optimisations

To implement fine-tuning, the following steps are applied:

1. **Pre-training:** Captures general domain-specific linguistic knowledge from the training data.
2. **Fine-tuning:**
   (a) Apply weighted loss functions to guide the LLMs into balancing performance with fairness.
   (b) Unfreeze lower layers to refine the representation of the stakeholder population slowly.
3. **Validation:** Test with evaluation metrics, such as Chi-Square, Cosine Similarity, and fairness metrics (DPD and EOD) to measure how the synthetic responses align with the real-world distributions.

### 3.3.4 Universal Language Model Fine-tuning Case Study

To further illustrate this process, an example fine-tuning method by Howard and Ruder uses universal language model fine-tuning (ULMFiT). This approach addresses overfitting on smaller training datasets and establishes a robust inductive learning structure for NLP tasks [11]. Specifically, this architecture uses three layers of Long Short-Term Memory (LSTM) networks, emphasizing the retention of conditioning knowledge, demonstrating a superior performance when compared with the existing text classification models.

Initially, a pre-trained language model trains on a general-domain corpus for capturing the general linguistic features. These pre-training steps provide the foundational knowledge for generalizing across diverse domains, such as environmental policymaking. In the second stage, fine-tuning optimises the model by performing some discriminative learning rates and slanted triangular learning algorithms on the target domain dataset. These techniques ensure that the LLMs can gradually adapt to task-specific features without overwriting foundational knowledge saved from the pre-trained state. Finally, an unfreezing operation adapts the lower layers of the model in tandem with the upper layers for preserving the robustness of the lower-level features.

While ULMFiT demonstrated superior performance in NLP tasks than other pre-training methods, it is more suitable for the LSTM-based models. In contrast, transformer-based models, such as GPT or BERT, use different fine-tuning techniques, like Adapters or LoRA (Low-Rank Adaptation). This methodology adapts the fine-tuning techniques with a hybrid approach that combines LoRA with sparse attention mechanisms. During fine-tuning, a dynamic parameter freezing strategy ensures that the relevant profiling parameters are updated.

## 3.4 Prompt Engineering

Prompt engineering involves crafting inputs that guide large language models (LLMs) to generate realistic and context-sensitive responses. This process includes two main types of prompts:

1. **System Prompts:** These establish the LLM's role and context by providing a detailed profile based on socio-demographic variables from the UKHLS datasets. For instance, a system prompt might specify an individual's age, income, and attitudes toward environmental issues (Fig. **2**).
2. **User Prompts:** These specify the task or question the LLM needs to address, helping it generate accurate and relevant predictions. For example, a user prompt might ask whether the individual is willing to pay more for environmentally friendly products (Fig. 3).

Ideologically, I describe myself as a Liberal Democrat supporter. Racially, I am British. I am male. My marital status is Single. In terms of my qualifications, My highest qualification is Secondary education. I have 5 children. I live in the Southeast. I live in a rural area. In terms of my age, my age group is 60-69 years old. My profession is Semi-Routine Occupations. When I asked to write my response to the question, "**And which of these would you say best describes your current lifestyle?**", I respond with **I do quite a few things that are environmentally friendly**.

Fig. 2: Example of a system prompt defining an individual's profile and their opinions about environmental issues.

Please answer this question "How strongly do you agree or disagree with the following statement: \'I would be prepared to pay more for environmentally friendly products.\'?" with one of the options without any additional explanation. Options:

1. Strongly Agree
2. Tend to Agree
3. Neither
4. Tend to Disagree
5. Strongly Disagree

Fig. 3: Example of a user prompt asking the LLM to determine willingness to pay for environmentally friendly products.

Both types of prompts could be customized for different demographics: younger individuals might be asked about participation in environmental organizations, while older groups might be asked about their willingness to adopt green technology. Tailoring prompts allowed the LLMs to simulate a broad range of public sentiments. Understanding public opinion depends on a set of interacting factors, which the prompts might oversimplify.

## 3.5 Evaluation Metrics

Several evaluation metrics assess how effective fine-tuning is at guiding the models into producing accurate public opinions:

1. **Chi-Square Tests**
2. **Cosine Similarity**
3. **Jaccard Index**
4. **Kullback-Leibler Divergence (KL Divergence)**

Chi-Square tests assess whether synthetic distributions align with the expected distributions from the UKHLS dataset, identifying any over- or under-representation from a particular group. Eq. 3 highlights how these tests help to evaluate the LLMs' ability to learn the relationship between demographic variables and public opinions on environmental policies. The Chi-Square test is crucial for objective 3, ensuring that the fine-tuned models' output distributions reflect the socio-demographic diversity like the real-world data, which is essential for accurately simulating public opinions on environmental policies.

$$\chi^2 = \sum \frac{(O_i - E_i)^2}{E_i} \quad (3)$$

Cosine similarity measures the similarities between synthetic and expected distributions, represented as vectors, with a range between -1 (dissimilar) and +1 (similar) (Eq. 4). It supports objectives 3 and 4 by highlighting the limitations at capturing similarities between categorical responses and representing the magnitude of differences between the synthetic and expected responses. Unlike other metrics, cosine similarity is a robust metric that remains unaffected by the scale of responses when comparing distributions, which helps account for minority or controversial opinions. As normalized values, cosine similarity measures the direction of opinion differences without being distorted by magnitude.

$$S_C(A, B) = \cos(\theta) = \frac{A \cdot B}{\|A\| \cdot \|B\|} = \frac{\sum_{i=1}^{N} A_i B_i}{\sqrt{\sum_{i=1}^{N} A_i^2} \sqrt{\sum_{i=1}^{N} B_i^2}} \quad (4)$$

When evaluating questions with categorical response options, particularly the questions with a binary set of response options, the Jaccard Index is great for quantifying the similarities and determining if a characteristic is present or absent from the synthetic responses. By converting the responses into a binary format, the Jaccard index divides the size of the intersection between the synthetic and expected distributions by the size of their union to produce a value between 0 (no overlap) and 1 (perfect overlap) (Eq. 5). The Jaccard Index supports objectives 1 and 3 by improving the models' ability to predict questions with a binary set of response options and evaluating how fine-tuned models could distinguish between different opinions.

$$J(A, B) = \frac{|A \cap B|}{|A \cup B|} \quad (5)$$

KL-divergence measures how much the synthetic distribution deviates from the expected distributions for the selected environmental questions (Eq. 6). If the KL-divergence score is 0, the synthetic and expected distributions match perfectly. As the KL-divergence score increases, the discrepancies between the distributions also grow. KL-divergence is effective at assessing probabilistic alignment between distributions, ensuring that the model generalizes across diverse socio-demographic groups. KL-divergence quantifies the discrepancies, helping to assess the limitations of fine-tuning and achieving objective 4.

$$D_{KL}(P||Q) = \sum_i P(i) \log \left( \frac{P(i)}{Q(i)} \right) \quad (6)$$

# 4 Results

## 4.1 Improvement through Fine-Tuning

Fine-tuning significantly improved the performance of large language models (LLMs) on ten questions related to attitudes toward environmental issues. Tables 3 and 4 compare the response distributions for the pre-trained versus fine-tuned LLMs across ten of the selected questions. Fig. A1 to A10 illustrate the alignment between synthetic and expected distributions achieved by fine-tuning LLMs to produce the expected responses. For example, Fig. A7 illustrates significant reductions from the overrepresentation of pro-environmental responses, aligning synthetic data with actual distributions.

## 4.2 Comparison of Pre-Trained vs. Fine-Tuned Models

Although the pre-trained models were powerful for performing general natural language processing (NLP) tasks, they struggle to interpret and generate domain-specific public opinions. Before fine-tuning, the models' output often fail to capture how different socio-demographic factors influence scepticism in environmental policymaking. An example of simulating public opinions on carbon taxes shows that a pre-trained model produces a uniform response distribution, failing to account for how variations in income levels or political party support influence individuals' opinions [2, 4].
After fine-tuning, the model produced more diverse responses by demographic groups. For example, when prompted with a profile of a 45-year-old and university-educated individual living in an urban area, the finetuned model predicted support for developing renewable energy infrastructure and increasing carbon taxes on businesses, an output mirroring responses from the UKHLS datasets. On the other hand, LLMs without finetuning produce oversimplified responses that ignore regional factors, assuming that people living in rural and urban areas share the same levels of support for renewable energy. Table 3 summarises a comparison between the pre-trained and fine-tuned models by policy support and shows more polarising levels of support for carbon taxes by income groups.

**Table 3** A comparison between the Pre-trained and Fine-tuned models in simulating public opinions on carbon taxes.

| Model | Support for Carbon Taxes (High Income) | Opposition to Carbon Taxes (Low Income) | Confidence Interval (±) |
|---|---|---|---|
| Pre-trained GPT-3.5 | 65% | 35% | ±5% |
| Fine-tuned GPT-3.5 | 82% | 55% | ±2% |

**Table 4** A table containing the synthetic response distributions to ten selected environmental questions generated by the LLMs, GPT-4o, GPT-4o mini, and GPT-4o1-preview, after the LLMs were fine-tuned.

| Question | Response Options | GPT-4o | GPT-4o mini | GPT-4o1-preview | Expected Distribution |
|---|---|---|---|---|---|
| **Describe your lifestyle** | Nothing | 5% | 6% | 4% | 5.74% |
| | Few | 32% | 36% | 33% | 35.66% |
| | Some | 40% | 37% | 42% | 40.45% |
| | Many | 21% | 17% | 20% | 16.23% |
| | All | 2% | 4% | 1% | 1.91% |
| **Climate Change Impact** | Entirely Positive | 2.11% | 3% | 4% | 2.13% |
| | More Positive than Negative | 10.53% | 15% | 16% | 10.52% |
| | Neither | 27.37% | 22% | 20% | 27.36% |
| | More Negative than Positive | 46.32% | 40% | 38% | 46.31% |
| | Entirely Negative | 13.68% | 20% | 22% | 13.68% |
| **Personal Impact on Climate** | Strongly Agree | 7% | 3% | 5% | 3.94% |
| | Tend to Agree | 8% | 7% | 11% | 8.62% |
| | Neither | 40% | 43% | 36% | 38.94% |
| | Tend to Disagree | 38% | 32% | 36% | 35.94% |
| | Strongly Disagree | 7% | 15% | 12% | 12.56% |
| **Willing to Pay** | Strongly Agree | 4% | 6% | 3% | 2.74% |
| | Tend to Agree | 9% | 7% | 8% | 6.94% |
| | Neither | 34% | 38% | 40% | 39.24% |
| | Tend to Disagree | 33% | 29% | 30% | 31.64% |
| | Strongly Disagree | 20% | 20% | 19% | 19.44% |
| **Personal Change** | Strongly Agree | 8% | 4% | 4% | 5.36% |
| | Tend to Agree | 4% | 3% | 7% | 5.92% |
| | Disagree | 53% | 49% | 55% | 53.21% |
| | Strongly Disagree | 31% | 36% | 30% | 33.14% |
| | Already Changed | 4% | 8% | 4% | 2.37% |
| **Environ. Disaster** | Strongly Agree | 21% | 19% | 17% | 17% |
| | Tend to Agree | 18% | 16% | 14% | 15.60% |
| | Neither | 35% | 36% | 35% | 34.50% |
| | Tend to Disagree | 26% | 24% | 25% | 25.50% |
| | Strongly Disagree | 0% | 5% | 9% | 7.40% |
| **Green Tariff** | Buy | 4% | 2% | 3% | 1.80% |
| | Considering | 20% | 18% | 21% | 19.80% |
| | No | 15% | 8% | 10% | 10.60% |
| | Rejected | 61% | 72% | 66% | 67.80% |
| **Pollution** | Yes | 7.80% | 11.50% | 6.50% | 3.49% |
| | No | 92.20% | 88.50% | 93.50% | 96.50% |
| **Environ. Group** | Mentioned | 30% | 32% | 28% | 50% |
| | Not Mentioned | 70% | 68% | 72% | 30% |
| **Climate Change Control** | Strongly Agree | 48% | 45% | 43% | 44.94% |
| | Tend to Agree | 1% | 3% | 2% | 1.45% |
| | Neither | 9% | 7% | 6% | 7.50% |
| | Tend to Disagree | 22% | 17% | 21% | 20.93% |
| | Strongly Disagree | 20% | 28% | 28% | 25.18% |

**Table 5** A table containing the synthetic response distributions to ten selected environmental questions generated by the pre-trained LLMs, GPT-4o, GPT-4o mini, and GPT-4o1-preview (without fine-tuning).

| Question | Response Options | GPT-4o (Pre-trained) | GPT-4o mini (Pre-trained) | GPT-4o1-preview (Pre-trained) | Expected Distribution |
|---|---|---|---|---|---|
| **Describe your lifestyle** | Nothing | 8% | 10% | 7% | 5.74% |
| | Few | 30% | 35% | 28% | 35.66% |
| | Some | 40% | 38% | 42% | 40.45% |
| | Many | 15% | 12% | 18% | 16.23% |
| | All | 7% | 5% | 5% | 1.91% |
| **Climate Change Impact** | Entirely Positive | 4% | 5% | 4% | 2.13% |
| | More Positive than Negative | 20% | 18% | 22% | 10.52% |
| | Neither | 25% | 30% | 24% | 27.36% |
| | More Negative than Positive | 40% | 37% | 42% | 46.31% |
| | Entirely Negative | 11% | 10% | 8% | 13.68% |
| **Personal Impact on Climate** | Strongly Agree | 18% | 20% | 17% | 3.94% |
| | Tend to Agree | 25% | 28% | 25% | 8.62% |
| | Neither | 30% | 25% | 32% | 38.94% |
| | Tend to Disagree | 17% | 18% | 18% | 35.94% |
| | Strongly Disagree | 10% | 9% | 8% | 12.56% |
| **Willing to Pay** | Strongly Agree | 12% | 10% | 14% | 2.74% |
| | Tend to Agree | 25% | 28% | 26% | 6.94% |
| | Neither | 30% | 32% | 30% | 39.24% |
| | Tend to Disagree | 20% | 22% | 20% | 31.64% |
| | Strongly Disagree | 13% | 8% | 10% | 19.44% |
| **Personal Change** | Strongly Agree | 20% | 22% | 18% | 5.36% |
| | Tend to Agree | 30% | 35% | 28% | 5.92% |
| | Disagree | 25% | 20% | 24% | 53.21% |
| | Strongly Disagree | 15% | 15% | 20% | 33.14% |
| | Already Changed | 10% | 8% | 10% | 2.37% |
| **Environ. Disaster** | Strongly Agree | 25% | 22% | 27% | 17% |
| | Tend to Agree | 30% | 32% | 28% | 15.60% |
| | Neither | 20% | 25% | 18% | 34.50% |
| | Tend to Disagree | 15% | 12% | 17% | 25.50% |
| | Strongly Disagree | 10% | 9% | 10% | 7.40% |
| **Green Tariff** | Buy | 15% | 12% | 18% | 1.80% |
| | Considering | 25% | 22% | 20% | 19.80% |
| | No | 50% | 53% | 49% | 10.60% |
| | Rejected | 10% | 13% | 13% | 67.80% |
| **Pollution** | Yes | 40% | 38% | 42% | 3.49% |
| | No | 60% | 62% | 58% | 96.50% |
| **Environ. Group** | Mentioned | 40% | 43% | 44% | 50% |
| | Not Mentioned | 60% | 57% | 56% | 30% |
| **Climate Change Control** | Strongly Agree | 12% | 10% | 14% | 44.94% |
| | Tend to Agree | 25% | 28% | 22% | 1.45% |
| | Neither | 30% | 25% | 32% | 7.50% |
| | Tend to Disagree | 20% | 22% | 20% | 20.93% |
| | Strongly Disagree | 13% | 15% | 12% | 25.18% |

### 4.3 Quantitative and Qualitative Analysis

Table 6 summarises the pre-trained and fine-tuned models' results from tables A1 and A2, which identified an improvement in the fine-tuned models' ability to include socio-demographic factors. Specifically, fine-tuning reduced the Chi-Square statistic by 21.5% (from 1.1827 to 0.9288) and KL-divergence scores by 30.3% (from 0.1630 to 0.1137). These optimisations indicate a notable improvement in the models' ability to align synthetic with actual responses, which are consistent with the improvements from the cosine similarity scores. Consequently, fine-tuned models could capture demographic and regional differences, such as the rural-urban divide in opinions on renewable energy infrastructure.

For example, fine-tuned models successfully captured nuanced opinions on carbon taxes, such as stronger support from higher-income urban populations, closely mirroring survey data. These results underline the practical utility of fine-tuned models in policy-making, enabling data-driven and inclusive decision-making that reflects demographic diversity. By accurately predicting public support for carbon taxes, fine-tuned models can help policymakers anticipate public reactions and refine legislative proposals with greater precision.

An average p-value of 0.3358 for fine-tuned models, compared to 0.2768 for pre-trained models, suggests that while differences between observed and expected distributions are not significant, fine-tuned LLMs are better equipped for mimicking public sentiments. These outcomes imply that policymakers can use these models to accurately predict public support for environmental initiatives, enabling more informed decision-making.

Cosine similarity scores revealed that fine-tuned models achieved a 4.08% increase (from 0.6792 to 0.72) in semantic alignment with expected distributions. Notably, this metric proved robust in capturing nuanced shifts in public sentiment from socio-demographic subgroups, such as varying support for renewable energy policies between urban and rural populations. However, its reliance on vector magnitudes suggests complementing it with categorical metrics, like Chi-Square, to ensure more comprehensive evaluations.

While the fine-tuned models achieved an average Jaccard Index of 0.72 compared to 0.6792 for pre-trained models, indicating improved accuracy in binary outputs, the improvement is less pronounced than for other metrics. For instance, responses to binary questions, such as "Are you living in a polluted area?" yielded marginal gains (0.7167 vs. 0.68). This suggests that binary-response questions require additional contextual features, such as regional air quality indices or localized environmental metrics, to capture subtle variations.

This limitation suggests that more synthetic examples are needed to highlight the subtle variation in the questions containing binary response options, such as more precise geographical data that could help determine the level of pollution alongside reports of local air quality index (AQI), Particulate Matter Concentration (PM2.5 and PM10), water quality index (WQI), and noise pollution levels (dB). These additional metrics create detailed profiling examples to understand how an individual answers binary-response questions. Both examples show a class imbalance in the pre-trained models with a higher proportion of people being members of an environmental organization and experiencing pollution (see Fig. A8 and A9).

**Table 6** Average performance scores of three pre-trained and fine-tuned LLM models, GPT-4o, GPT-4o-mini, and GPT-4o-previews, for the metrics, Chi-Square test, Cosine Similarity, Jaccard Index, and KL-Divergence

| ***Model*** | Chi-Square (**Fine-tuned**) | Chi-Square (**Pre-trained**) | Cosine Similarity (**Fine-tuned**) | Cosine Similarity (**Pre-trained**) |
|---|---|---|---|---|
| **GPT-4o** | 0.8678 | 1.1600 | 0.9322 | 0.9000 |
| **GPT-4o mini** | 0.9911 | 1.2000 | 0.9422 | 0.9056 |
| **GPT-4o1-preview** | 0.9244 | 1.1881 | 0.9389 | 0.9044 |
| **Average** | **0.9288** | **1.1827** | **0.9378** | **0.9033** |
| ***Model*** | Jaccard Index (**Fine-tuned**) | Jaccard Index (**Pre-trained**) | KL-divergence (**Fine-tuned**) | KL-divergence (**Pre-trained**) |
| **GPT-4o** | 0.7200 | 0.6778 | 0.1211 | 0.1700 |
| **GPT-4o mini** | 0.7256 | 0.6811 | 0.1022 | 0.1622 |
| **GPT-4o1-preview** | 0.7144 | 0.6789 | 0.1178 | 0.1567 |
| **Average** | **0.7200** | **0.6792** | **0.1137** | **0.1630** |

# 5 Discussion

## 5.1 Fine-Tuning as a Key Element in LLM Design

Fine-tuning general-purpose large language models (LLMs) enables them to tackle domain-specific tasks with higher accuracy and relevance, particularly in fields like environmental and climate sciences (Objective 3). This study demonstrates that fine-tuning leads to significant improvements in model performance, such as a reduction in Chi-Square and KL-divergence scores, which show better alignment with real-world data from the UKHLS datasets (Objective 1). These findings align with the previous research [24], which optimized fine-tuned models to achieve better performance without compromising accuracy.

This study demonstrates that fine-tuned models addressed the overrepresentation of individuals highly aware of their environmental impact, a significant bias observed in pre-trained models (see Fig. A3). To mitigate the biases, gradient accumulation ensures that larger datasets can be split into smaller batches, where they are processed over multiple iterations to build up the gradient value before updating weight parameters. This update helps developers with classifying views from minority groups (see Fig. A1 to A10).

Building on this foundation, including tailored profiling variables, such as job insecurity and spending habits, revealed additional factors influencing renewable energy spending decisions [25]. These profiling variables enhanced the fine-tuned models' ability to identify stakeholders who are against spending more on green tariff electricity (see Fig. A4). However, introducing profiling variables, such as income and spending habits, raises ethical concerns regarding reinforcing existing stereotypes about certain demographics (see Table 2). While fine-tuning enhances performance, it also introduces the potential for perpetuating societal biases, making it essential for researchers to balance model efficacy with fairness. Building on the technical benefits of fine-tuning, it is crucial to examine the broader societal and ethical implications of this approach.

## 5.2 Challenges in Fine-Tuning

One significant challenge in fine-tuning LLMs is overfitting, especially when working with smaller or more domain-specific datasets such as the UK Household Longitudinal Study (UKHLS) datasets (Objective 4). The UKHLS dataset, collected between January 2009 and May 2023, has a temporal gap that might not capture more recent technological developments, awareness of climate change, and shifts in the geo-political landscape. Therefore, using the UKHLS as reference data might amplify existing biases between the dataset and contemporary opinions, which mistakenly guides the LLMs in the wrong direction for fine-tuning.

To mitigate the consequences of these challenges, it is vital to ensure that the training datasets are not only representative but also dynamically updated to reflect current societal shifts. Further data collection ensures that the fine-tuned models evolve in parallel with public opinions instead of reinforcing outdated narratives.

When selecting the data, sampling bias remains a critical issue when providing accurate training datasets for conditioning and comparing the LLM-generated synthetic responses with (Objective 4). In the conditioning dataset, there are missing and invalid profile parameters, such as 97% of the "qfhigh" (highest qualification) variable being inapplicable (-8). After imputing invalid values for the highest qualification obtained variable, the distribution might not be nationally representative. For example, comparing the selected UKHLS variable with the YouGov Survey results found conflicting proportions of people with university degrees (UKHLS 7.36% versus YouGov 50.9%). If the expected data distribution does not represent the population, it could cause the developers to make incorrect conclusions about the accuracy of their synthetic data and make unnecessary fine-tuning steps. For example, if the developers overrepresent the proportion of people with higher education qualifications, the models might contribute to more environmental-friendly initiatives, like green tariffs and carbon taxes, which might be counterproductive and undermine support for green policies.

## 5.3 Ethical Considerations

Fine-tuning LLMs have profound ethical implications. For example, the LLMs inherit biases from their training data, making it difficult to select a perfect sample to train the models. While the data typically comes from surveys designed to be representative at national or regional levels, these surveys often face significant non-response issues from certain socio-demographic groups. Reducing bias requires precise data processing, such as using tailored questionnaires to track stakeholder interests. These trade-offs between accuracy and fairness raise questions about how practical are the LLMs' applications in the real-world environment. For example, in domains like environmental policymaking, less accurate models could undermine the people's trust in AI systems if they fail to accurately predict consensus views. Conversely, systematic biases could occur and marginalize the underrepresented communities.

Another concern is about the transparency of the LLMs and how they generate the outputs based on the roles defined through the profiling variables. To ensure transparency, it is vital to record profiles used to condition the LLMs. Having a transparent model is vital for building trust with stakeholders, so that they know how their data is used. Counterfactuals are great for learning about different opinions as observed in the two examples from Table 7 that shows two opposite opinions about whether climate change is controllable. These ethical and technical challenges underscore the need for robust strategies to optimize fine-tuning and guide future research directions.

**Table 7** A table shows two examples of counterfactual profiles about how an individual answers a question about whether climate change is controllable

| | |
|---|---|
| Ideologically, I describe myself as a Green Party supporter. Racially, I am Mixed-race. I am a female. My marital status is Single. In terms of my qualifications, my highest qualification is bachelor's degree. I have no children. I live in the Southwest. I live in a rural area. In terms of my age, my age group is 25-29 years old. My profession is Creative Occupations. When asked to write my response to the question, "And which of these would you say best describes your current lifestyle?", I respond with I do a lot of things that are environmentally friendly. I **strongly agree** that climate change is controllable. | (a) |
| Ideologically, I describe myself as a Reform UK supporter. Racially, I am White British. I am male. My marital status is Married. In terms of my qualifications, my highest qualification is Secondary education. I have 3 children. I live in the Northeast. I live in a suburban area. In terms of my age, my age group is 45-49 years old. My profession is Skilled Trades. When asked to write my response to the question, "And which of these would you say best describes your current lifestyle?", I respond with I do some things that are environmentally friendly. I **tend to disagree** that climate change is controllable. | (b) |

## 5.4 Evaluation Metrics and Their Limitations

In this study, several evaluation metrics were applied to assess the fine-tuned models, each offering unique insights into the models' performances. However, it is important to acknowledge the limitations present from each of these metrics.

For instance, Cosine Similarity is effective in comparing the direction of opinions, it does not account for the magnitude of differences in responses, potentially overlooking important variations, such as opinion intensity. Most cosine similarity scores fall within a narrow range (0.87 – 0.98), which restricts their ability at distinguishing between similar distributions. Both fine-tuned (0.91 – 0.98) and pre-trained models (0.87 – 0.93) achieved a similar range of cosine similarity scores, despite having different distribution shapes. This metric alone could potentially misrepresent how models emulate public sentiment, especially for polarising questions about environmental issues.

KL-Divergence is sensitive to minor differences from using high-dimensional data, which might not always be meaningful, especially in the sparse responses. However, when comparing the pre-trained with the fine-tuned distributions in Fig. A1, both models achieved similar KL-divergence scores due to similar overall trends from both datasets.

Chi-Square tests, though useful for evaluating alignment between the synthetic and expected distributions, can be prone to errors if the expected distributions are misrepresented. The Jaccard Index, while beneficial for binary responses, might fail to capture complex or nuanced opinions when applied to a larger number of categories. Despite Fig. A10 showed improvement in distribution alignment when moving from pre-trained to fine-tuned models, the Jaccard index did not reflect this enhancement significantly (Pre-trained score of 0.67 versus a Fine-tuned score of 0.74). Combining these metrics offers a holistic evaluation of the models' performance. Using multiple approaches, this study mitigates the limitations of each metric, ensuring a comprehensive assessment of how fine-tuned models can simulate opinions on environmental policies from a broad scope of perspectives.

# 6 Future Work

## 6.1 Multi-Task Learning for Optimized Fine-Tuning

While this study demonstrates how effective fine-tuning is at enhancing LLMs for generating public opinions, there are areas for exploring how multi-tasking learning optimises fine-tuning by allowing the LLMs to process multiple tasks simultaneously while aiming to reduce the risks of overfitting. Transfer learning allows LLMs to multi-task by processing the training data subsets efficiently, especially when the data is sparse or expensive to obtain, such as in environmental policy simulations. For example, Pilaut et al. (2020) identified a method for multi-task learning to help select relevant profiling parameters for conditioning LLMs [26]. This method balanced the weighting of the parameters and achieved 2.2% performance improvement compared to fully finetuned BERT models when applied to the GLUE benchmark, while compressing the data to 64.6% of its original volume. Applying similar techniques could enable simultaneous training on diverse profiling variables, reducing computational overhead by at least 30% for environmental policy simulations.

## 6.2 Adaptive Fine-Tuning and Layer Optimization

Exploring adaptive fine-tuning methods, such as the LRBench framework proposed by Jin et al. (2023), can help identify optimal learning rates and training depths for different LLM layers. Unlike uniform fine-tuning approaches, layer-wise optimization ensures that the high-relevance layers could maintain their focus without compromising general knowledge retention [27]. Preliminary studies indicate that adaptive fine-tuning can improve task-specific performance by up to 15%. For environmental applications, this technique could refine LLMs' understanding of variables, such as public attitudes toward region-specific renewable energy policies.

## 6.3 Active Learning to Address Data Sparsity

Integrating active learning into LLMs ensures that they can query informative samples for learning during fine-tuning. For instance, Mahalingam et al. (2023) demonstrated that active learning reduced the number of labelled data required by up to 40%, while improving classification accuracy in image recognition tasks [28]. Adapting this method to policy simulations could help mitigate issues from underrepresented demographic groups, such as the rural or lower-income populations. Active learning can also ensure dynamic model updates, aligning LLMs with evolving public opinions and thereby minimizing biases.

# 7 Conclusion

## 7.1 Summary of Findings

This study has demonstrated that fine-tuning prepares large language models (LLMs) for domain-specific applications, such as simulating opinions for reshaping environmental policies. After fine-tuning, models such as GPT-4o, GPT-4o-mini, and GPT-4o1-preview, produced improved synthetic responses that closely align with real-world distributions, as evidenced by the Chi-Square test scores, Cosine Similarity, KL-divergence scores, and Jaccard Indexes (see Table 5). These metrics allow the LLMs to account for socio-demographic factors, like regional, income, and educational differences. As a result, pre-trained models tend to generalize, producing more opinions favouring environmental policies. Fine-tuning improved the response distribution and allowed the representation of minority opinions, such as determining if an individual is willing to pay for green products and whether an individual wants to join an environmental organisation (see Fig. A1 – A10). These results affirm that fine-tuning is essential to transform LLMs from general-purpose tools into specialized systems that can simulate complex public opinions accurately.

## 7.2 Broader Implications

Fine-tuning has emerged as a foundational tool to prepare LLMs to meet its domain-specific requirements. Recent applications include integrating fine-tuned models for detecting hallucinations in machine-translated text, which showed excellent performances across different languages [29], and simulating how the American population reacts to diverse stimuli, achieving prominent levels of correlation between synthetic and real-world test data [30]. This study contributes to the growing evidence that fine-tuning improved the outputs' relevance. In high-stakes fields, such as healthcare, education, law, and finance, where precision and accountability are vital, fine-tuning ensures dependable and domain-specific outputs.

Furthermore, the success of fine-tuned models in simulating opinions on environmental policies underscores their potential for reshaping existing policies and boosting public engagement with scalable tools for learning diverse societal attitudes. As the demand for LLMs grow, fine-tuning will be vital for guiding the models into meeting the ethical and technical guidelines. Embedding data privacy into RoBERTa model helped them to achieve an 87.8% accuracy with a privacy budget of $\varepsilon = 6.7$, which confirms the maintenance of high accuracy [31]. The future developments will focus on reducing the computational costs of fine-tuning while enhancing the LLMs' capabilities across multiple domains.

In conclusion, this research defines the indispensable role of fine-tuning in optimising LLMs. The results set the stage for performing further explorations into how to select suitable conditioning datasets for implementing more dynamic domain-specific multi-tasking missions.

# Appendix

## Appendix 1: Figures

(a)

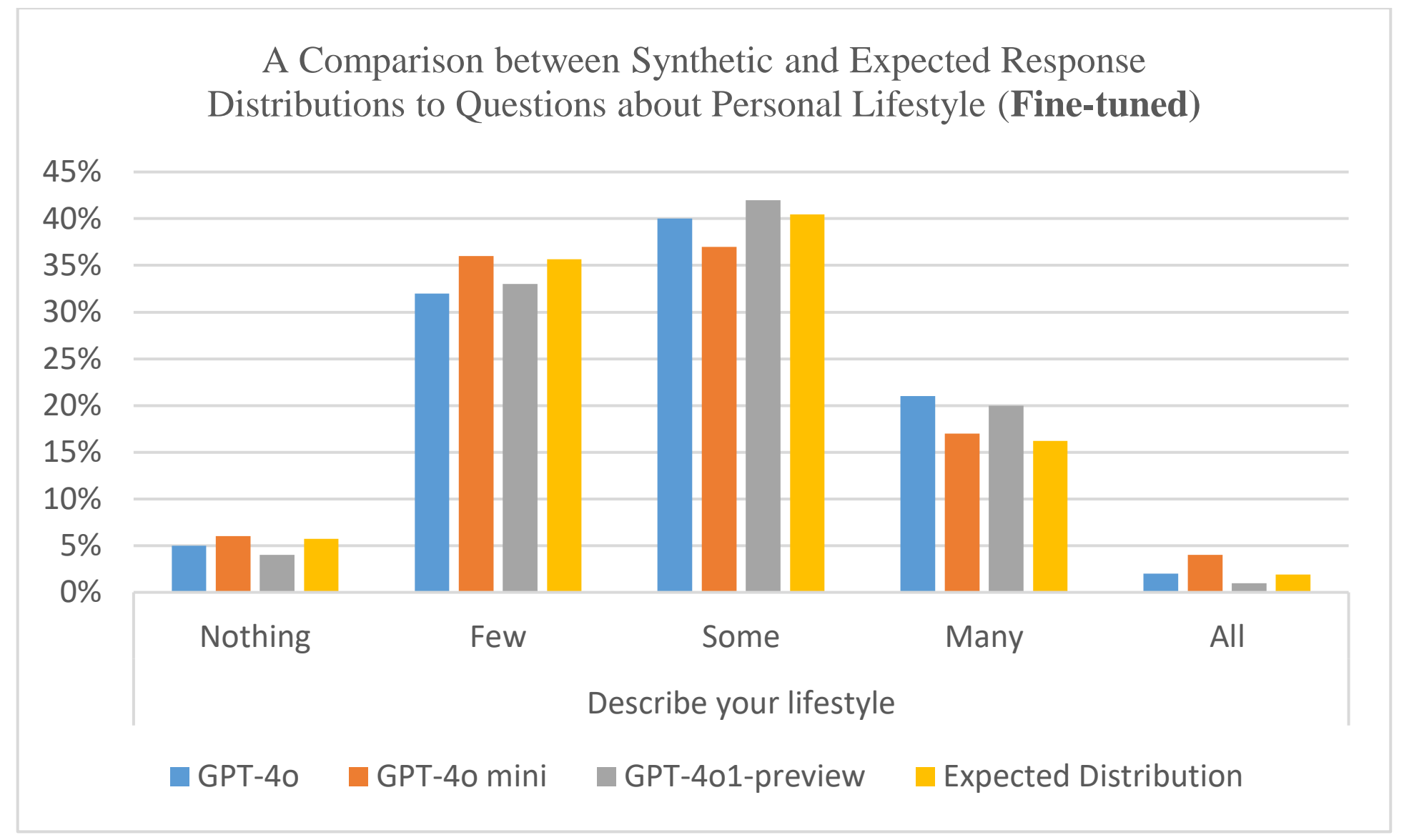

(b)

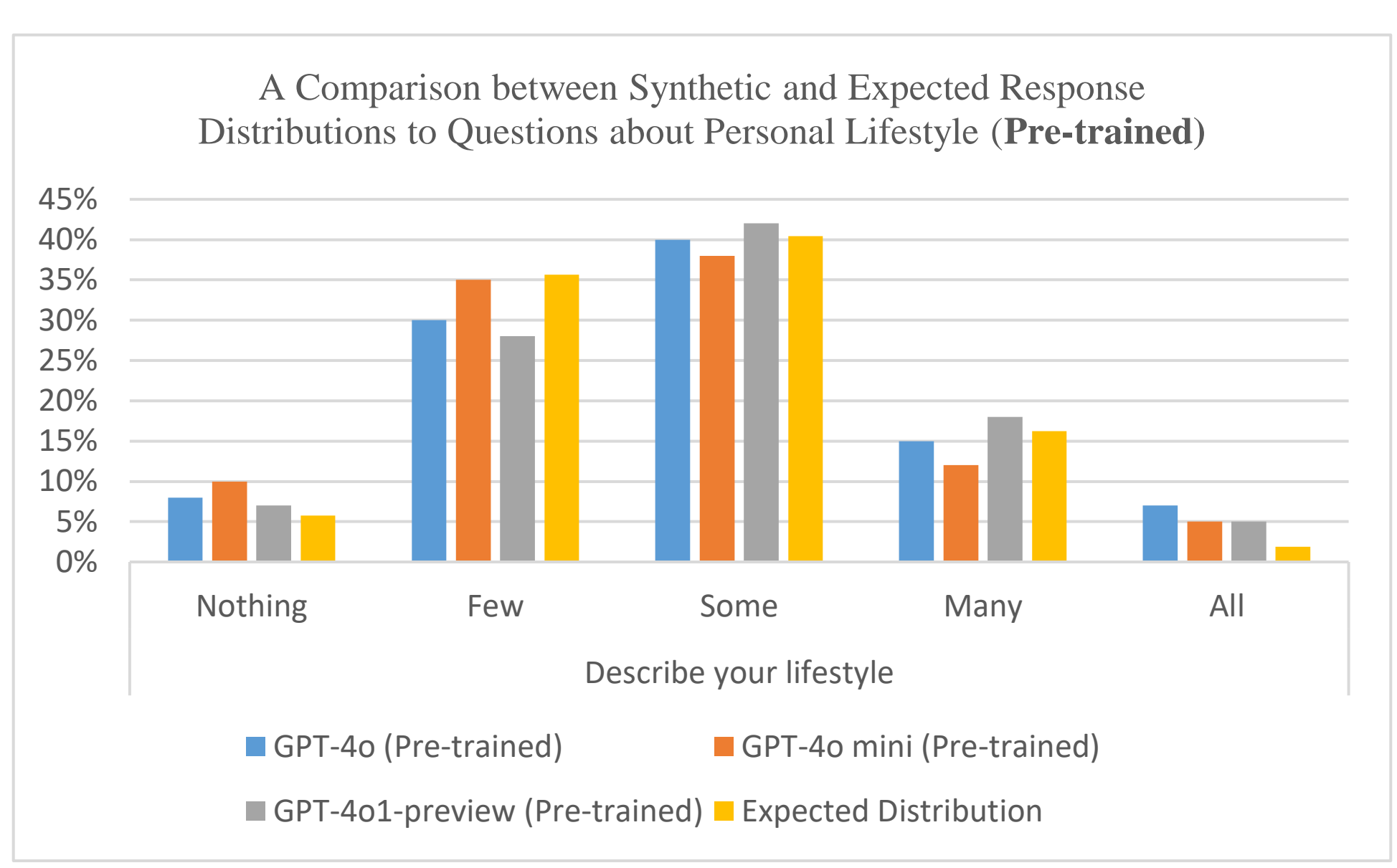

**Fig. A1** Graphs comparing the synthetic and expected response distributions to a question asking if an individual's lifestyle is environmentally friendly: (a) fine-tuned (b) without fine-tuning.

(a)

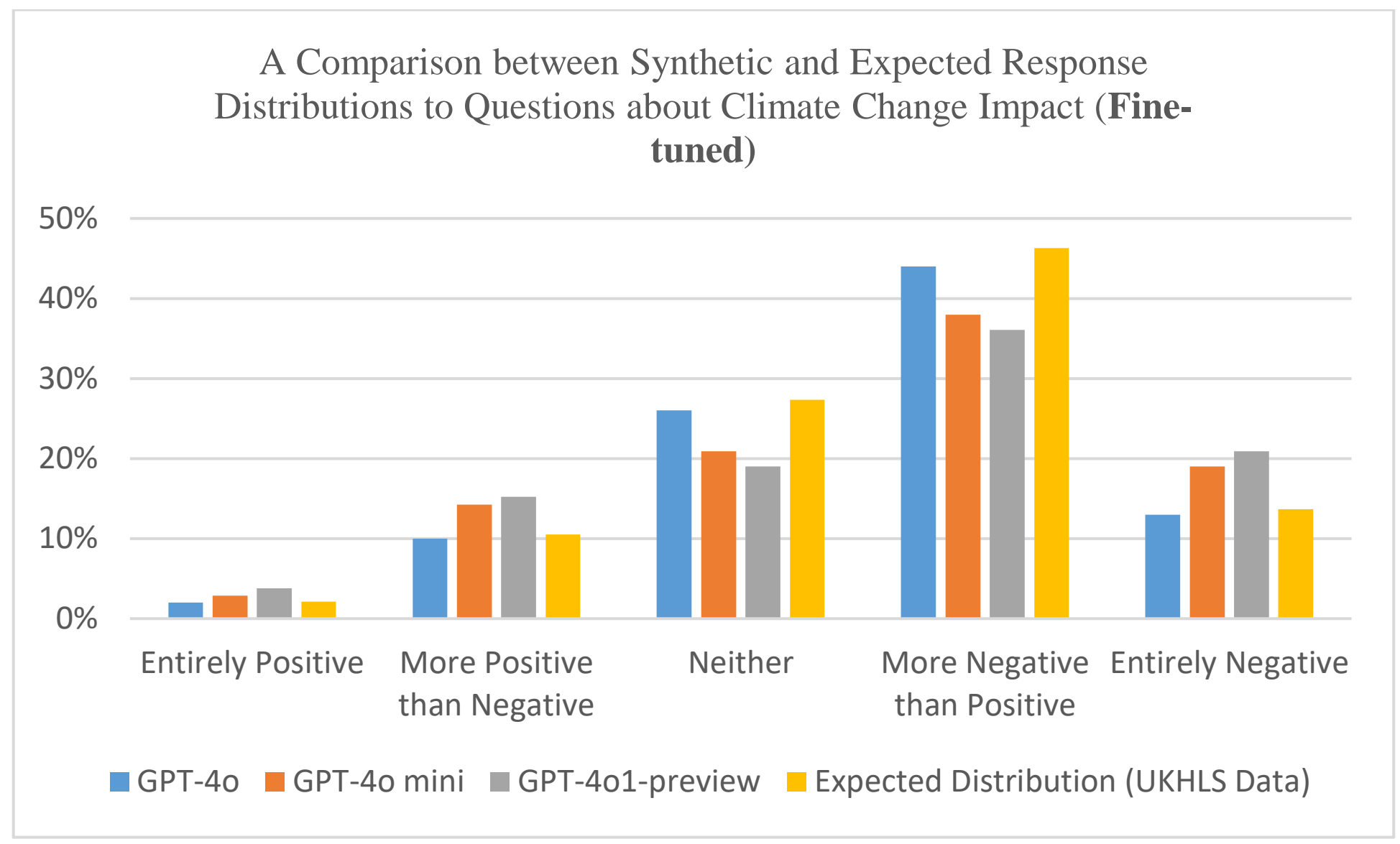

(b)

**Fig. A2** Graphs comparing the synthetic and expected response distributions to a question asking about how an individual perceives the impact of climate change.

(a)

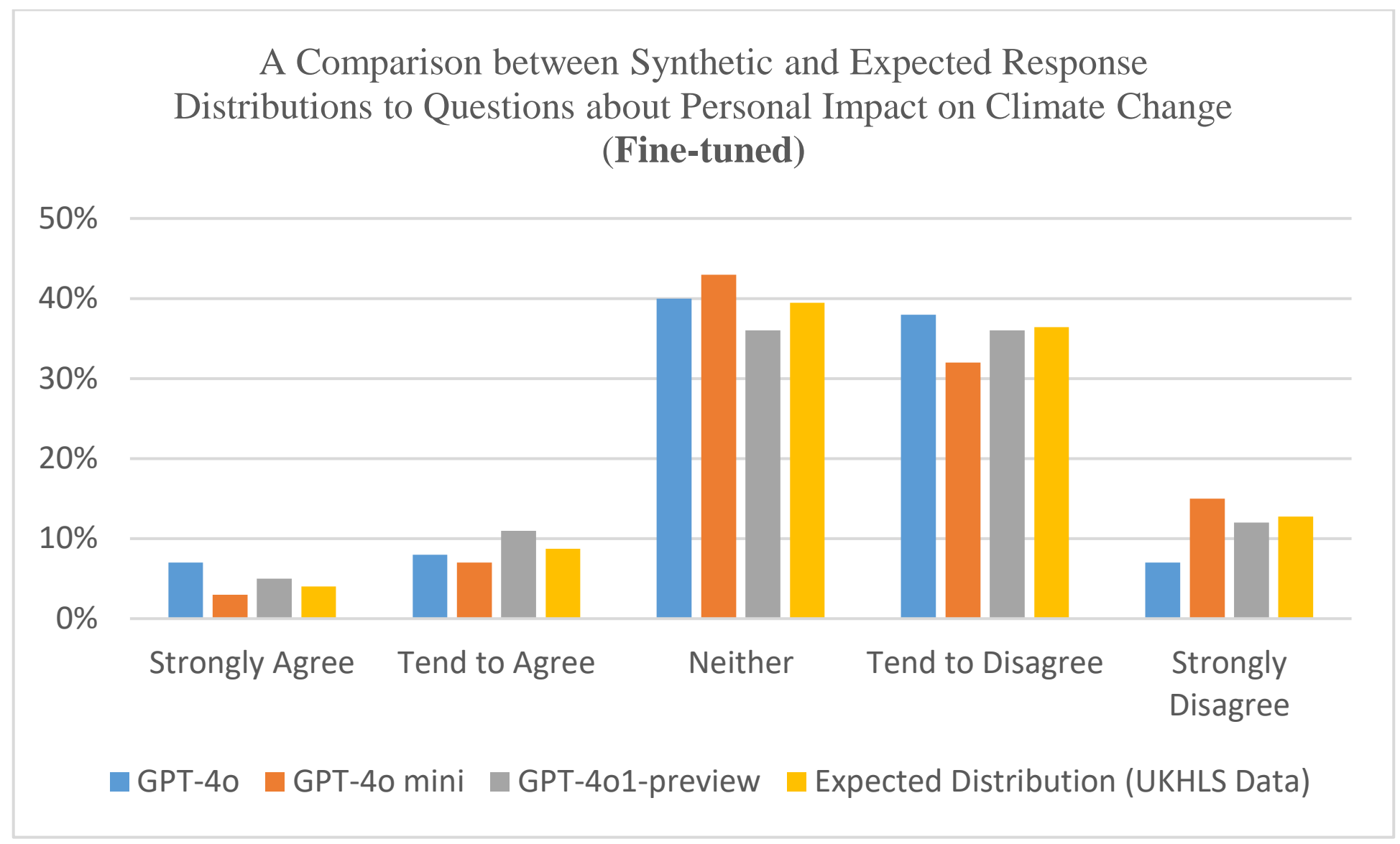

(b)

**Fig. A3** Graphs comparing the synthetic and expected response distributions to a question asking about how an individual perceives the personal impacts of climate change: (a) fine-tuned (b) without fine-tuning.

(a)

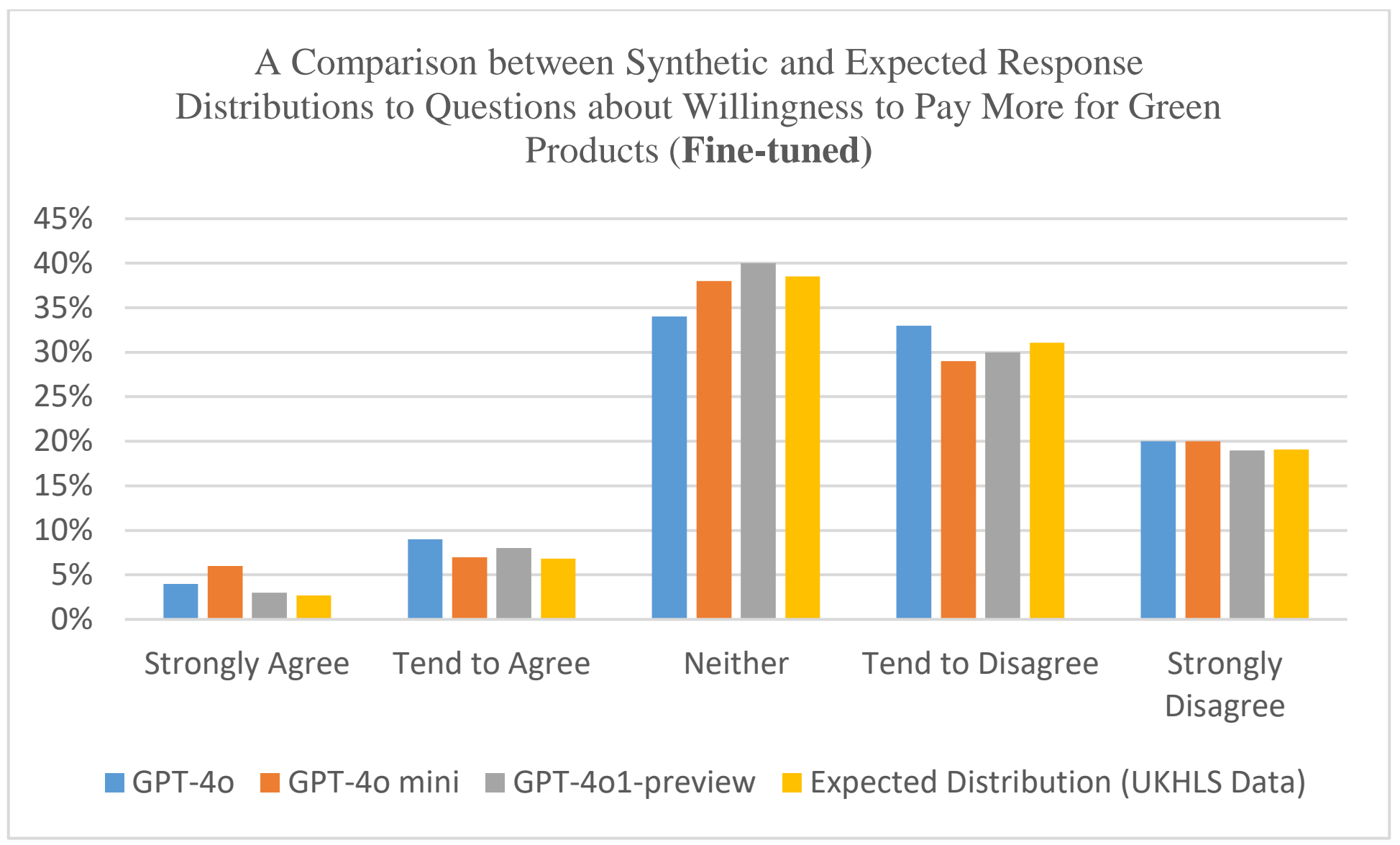

(b)

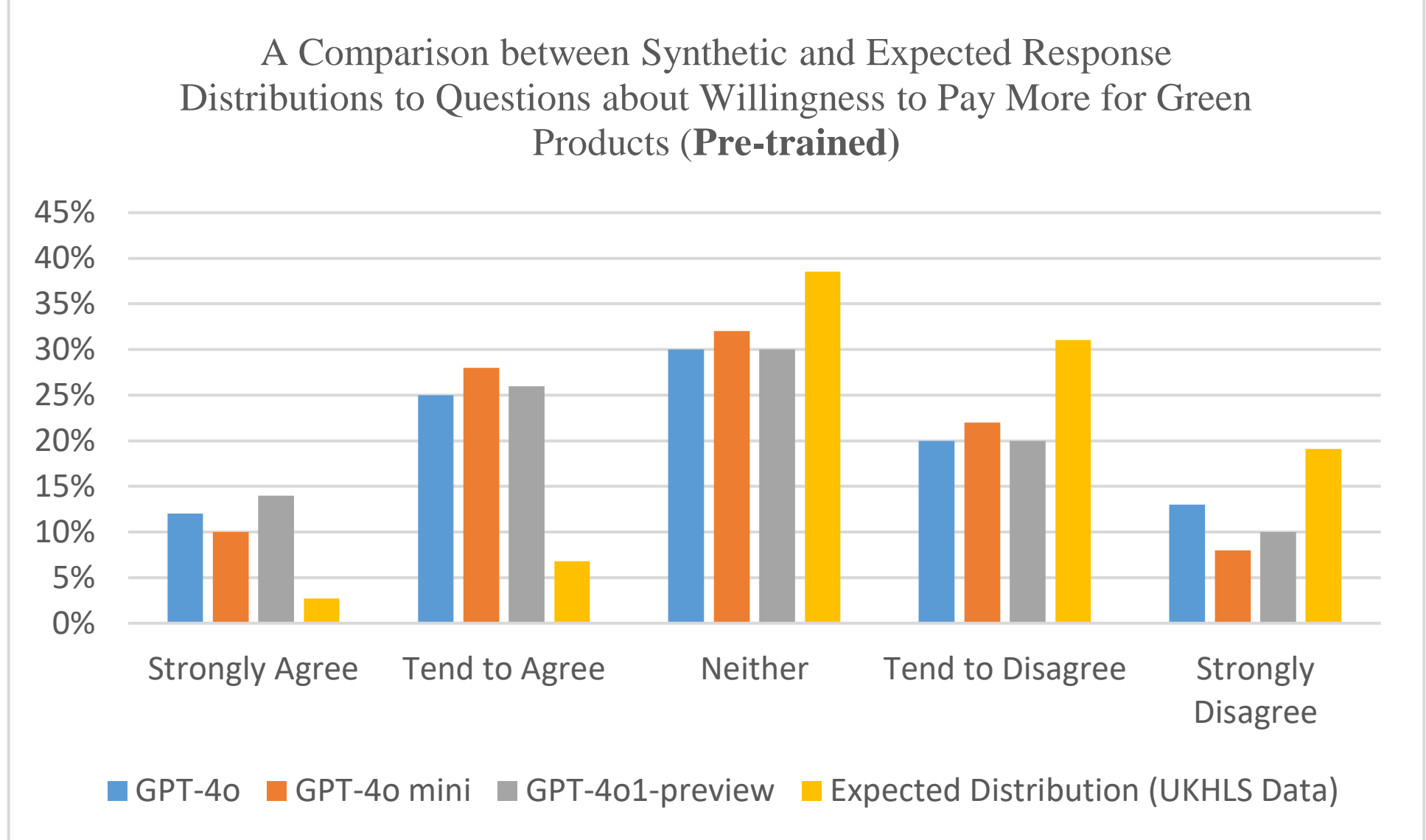

**Fig. A4** Graphs comparing between the synthetic and expected response distributions to a question asking about if an individual is willing to pay more for green products.

(a)

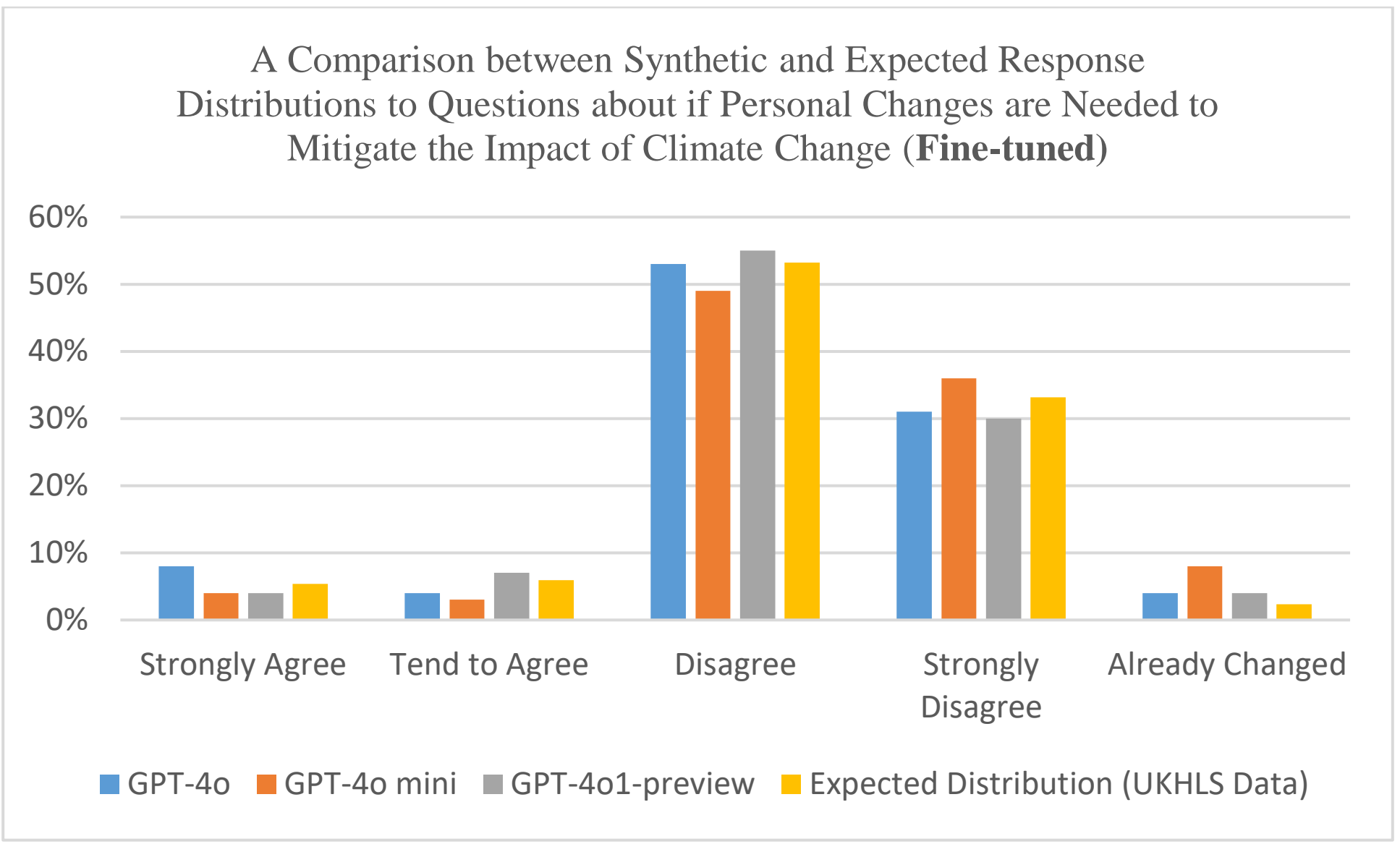

(b)

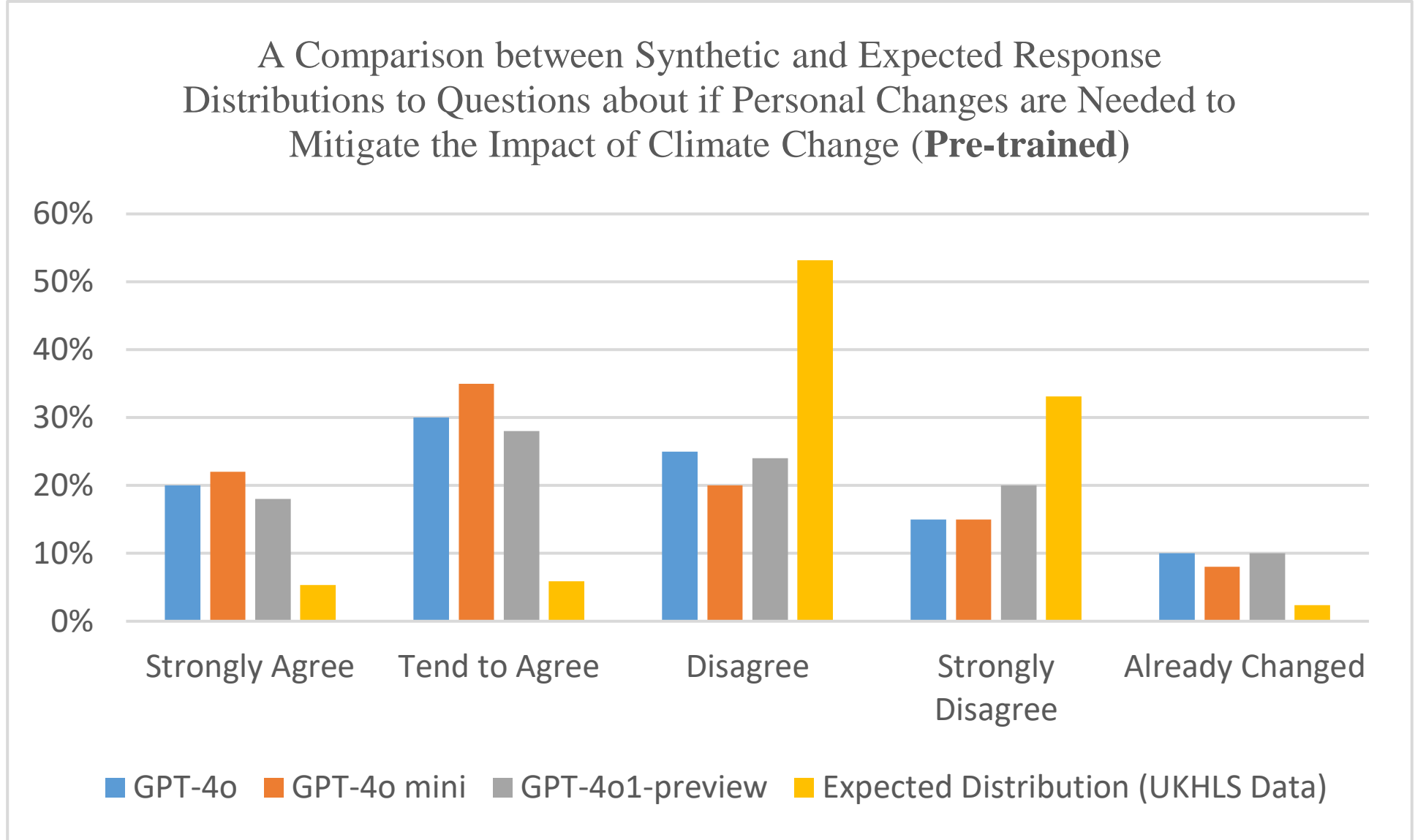

**Fig. A5** Graphs comparing the synthetic and expected response distributions to a question asking about if an individual needs to change their habits to mitigate the impact of climate change.

(a)

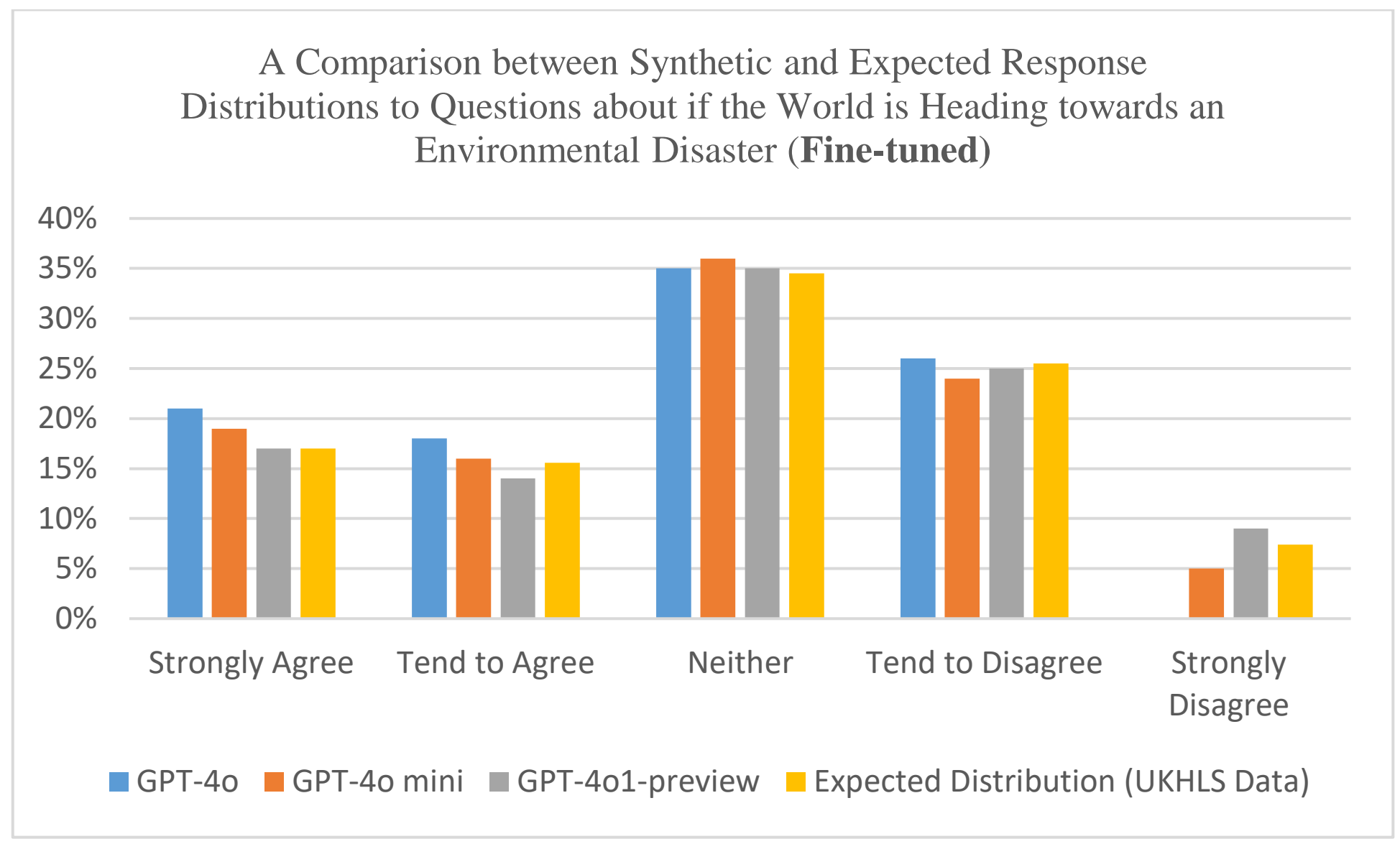

(b)

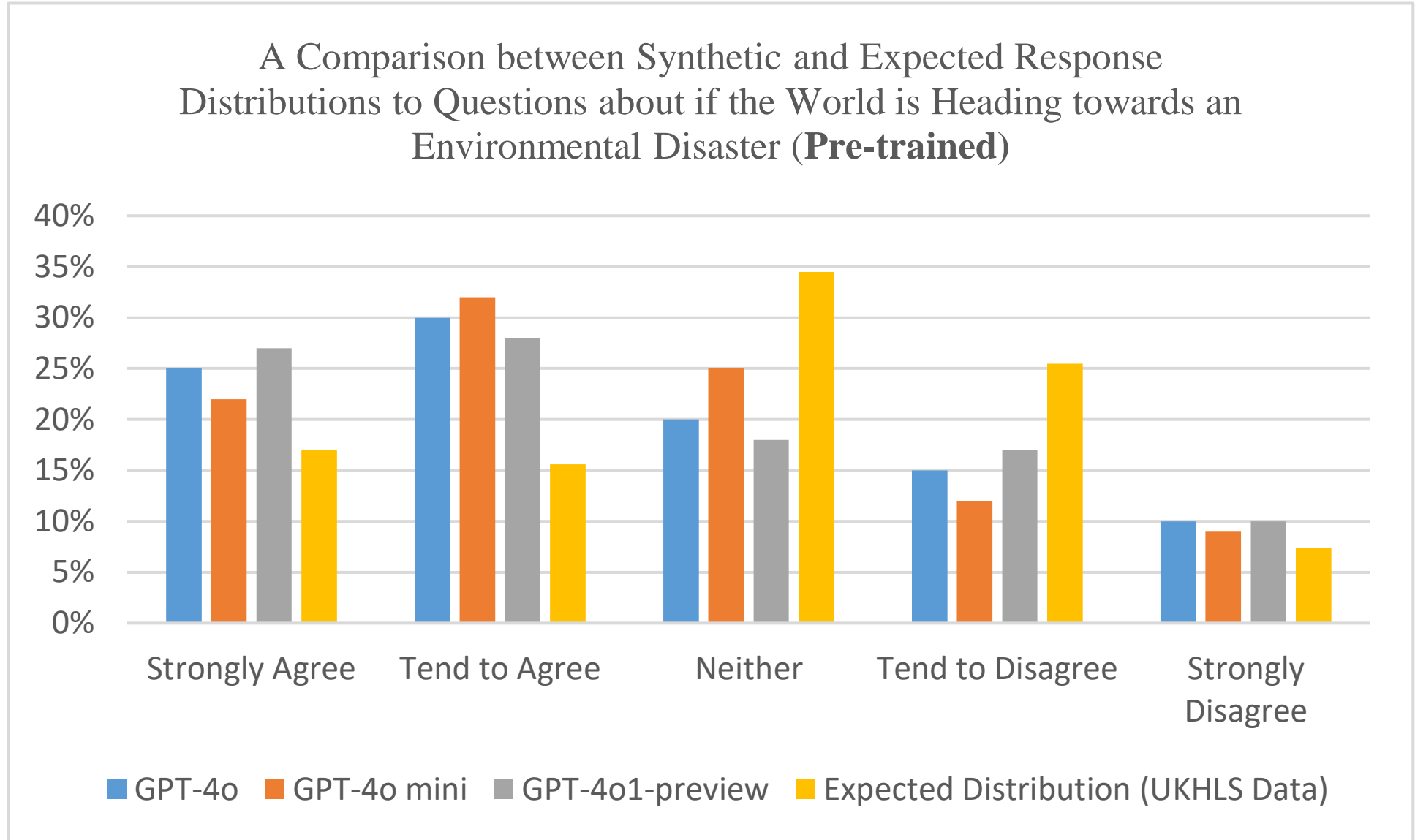

**Fig. A6** Graphs comparing the synthetic and expected response distributions to a question asking about if an individual thinks that the world is heading toward environmental disasters if people do not change their environmental behaviours.

(a)

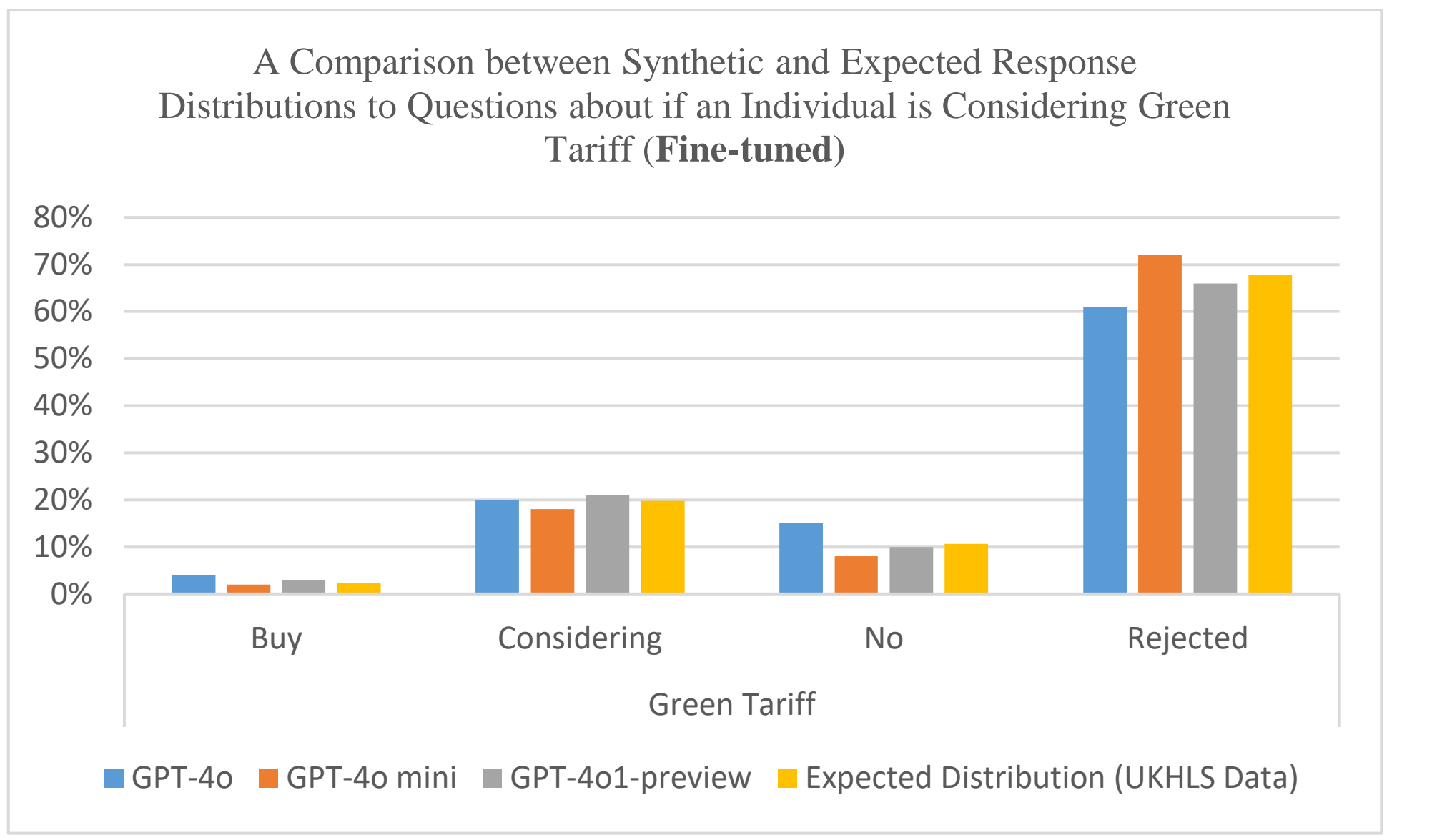

(b)

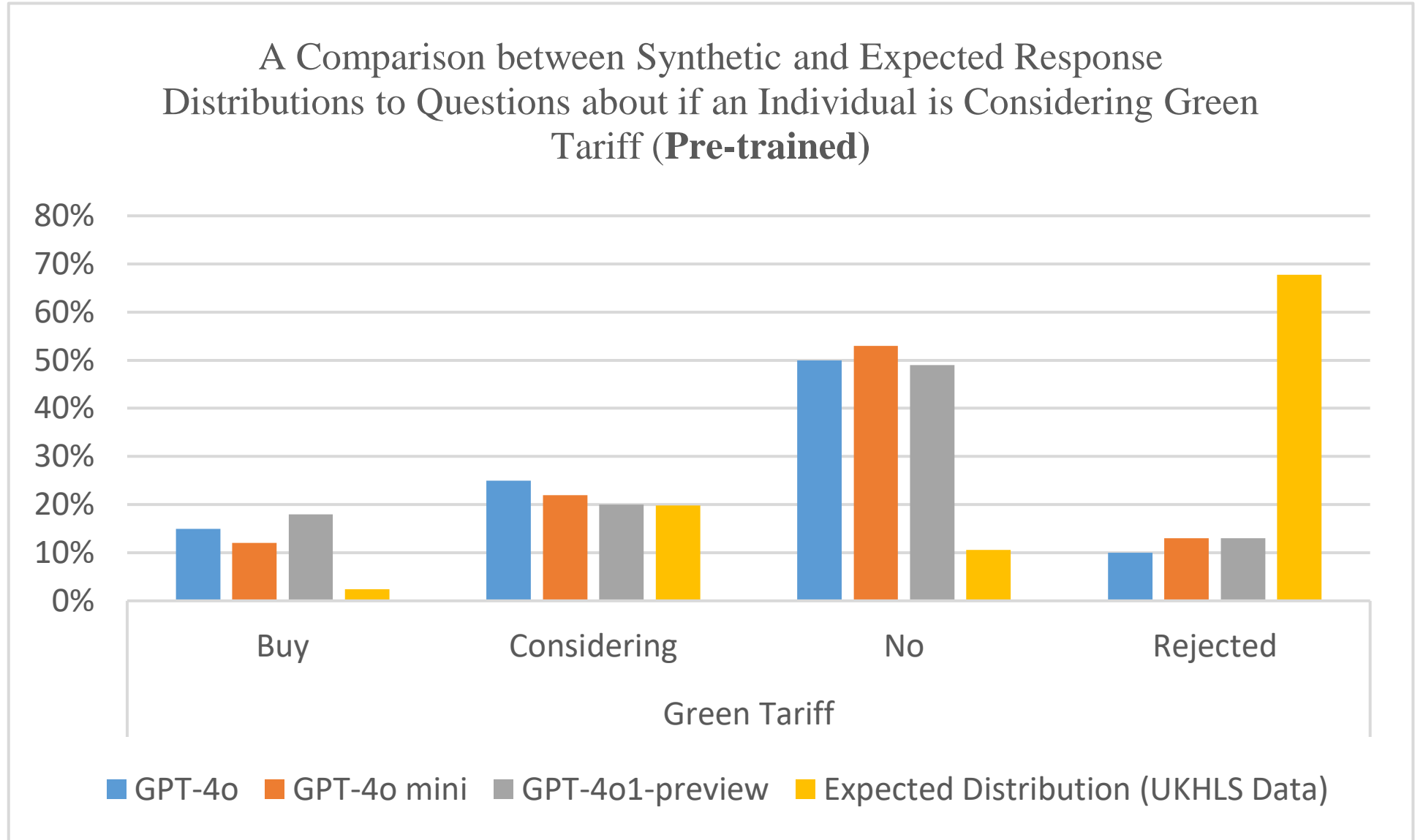

**Fig. A7** Graphs comparing the synthetic and expected response distributions to a question asking about if an individual is considering buying green tariff as an alternative source of electricity or gas supply.

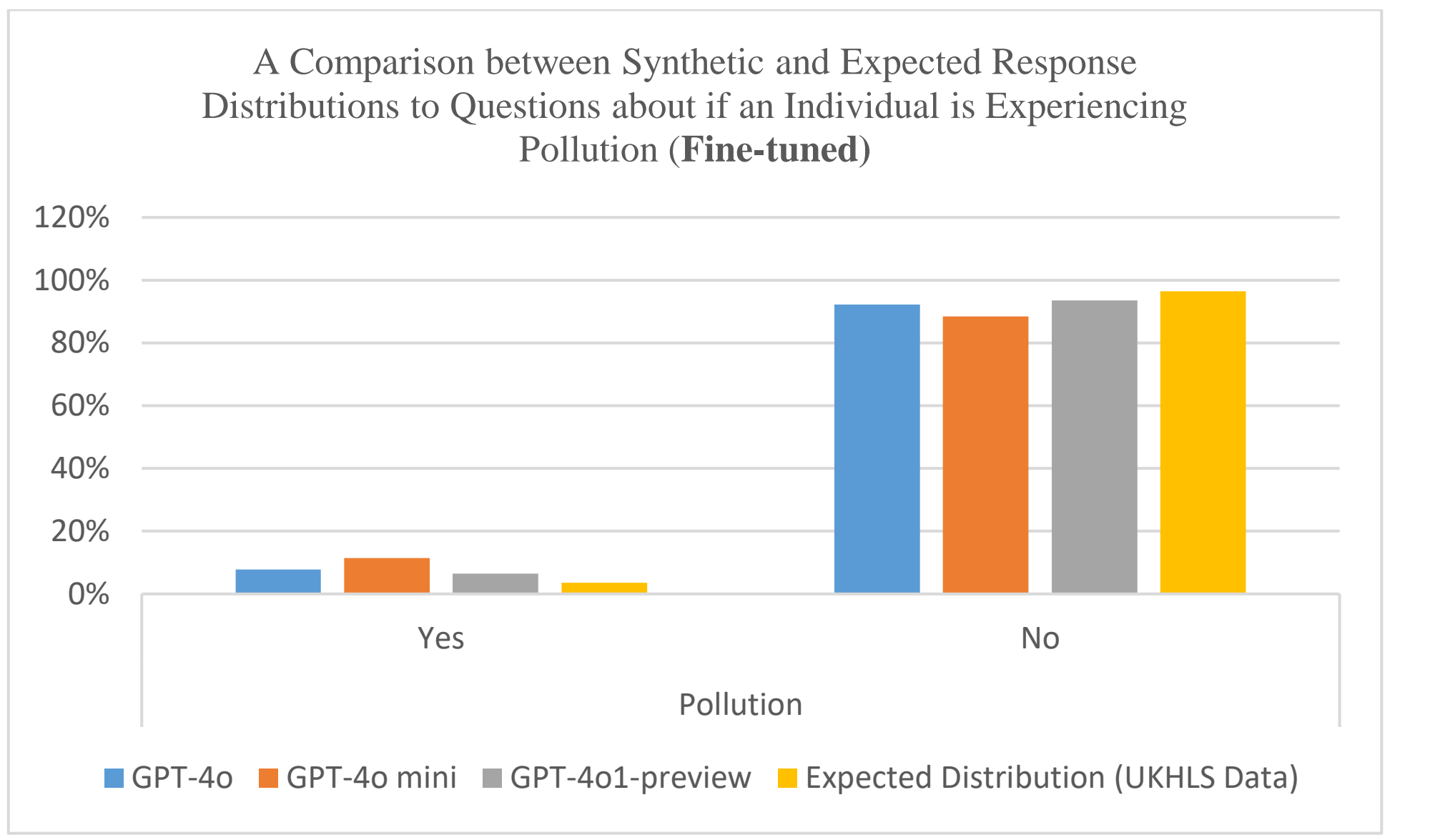

(a)

(b)

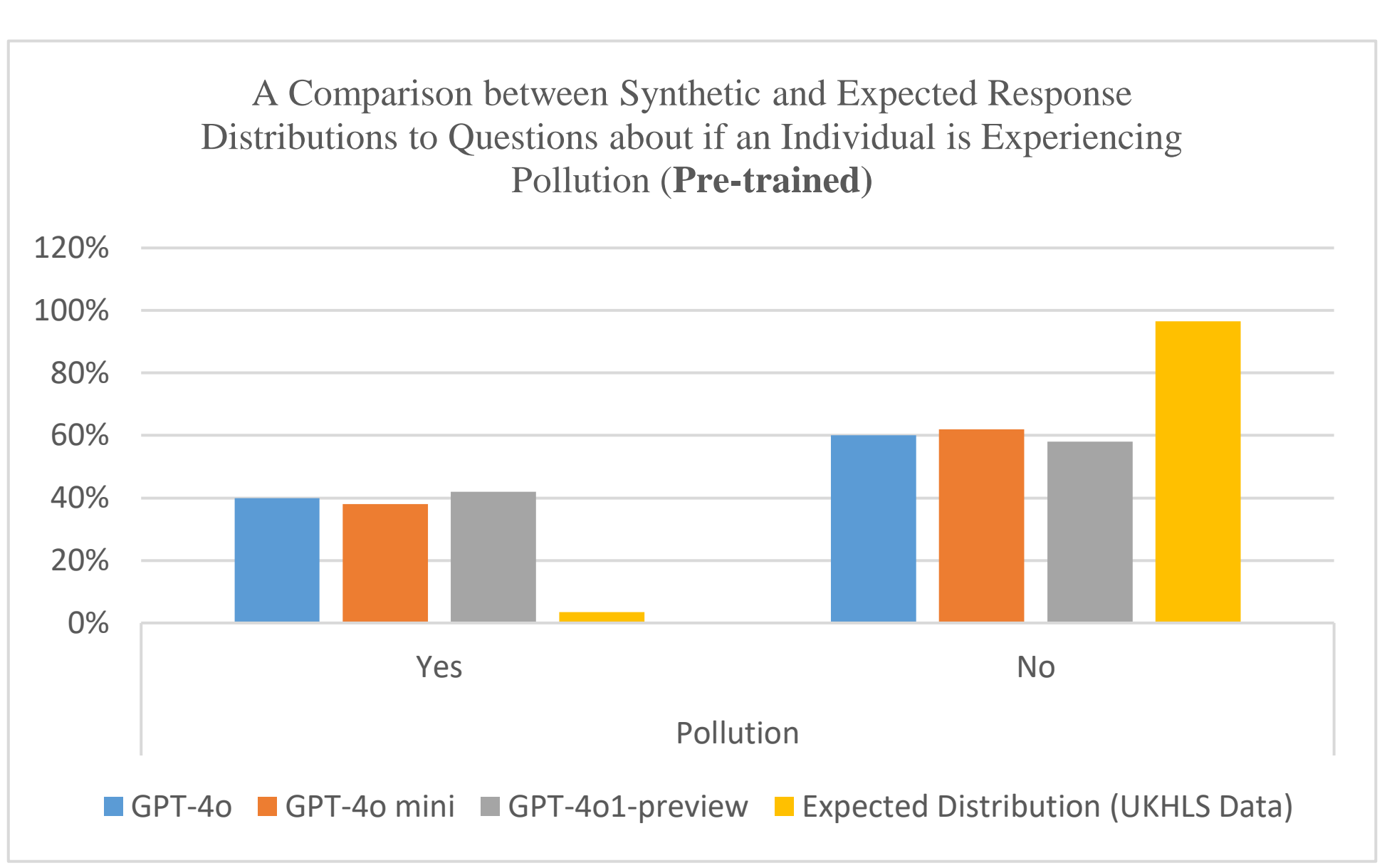

**Fig. A8** Graphs comparing the synthetic and expected response distributions to a question asking about if an individual is living in an area that experiences pollution.

(a)

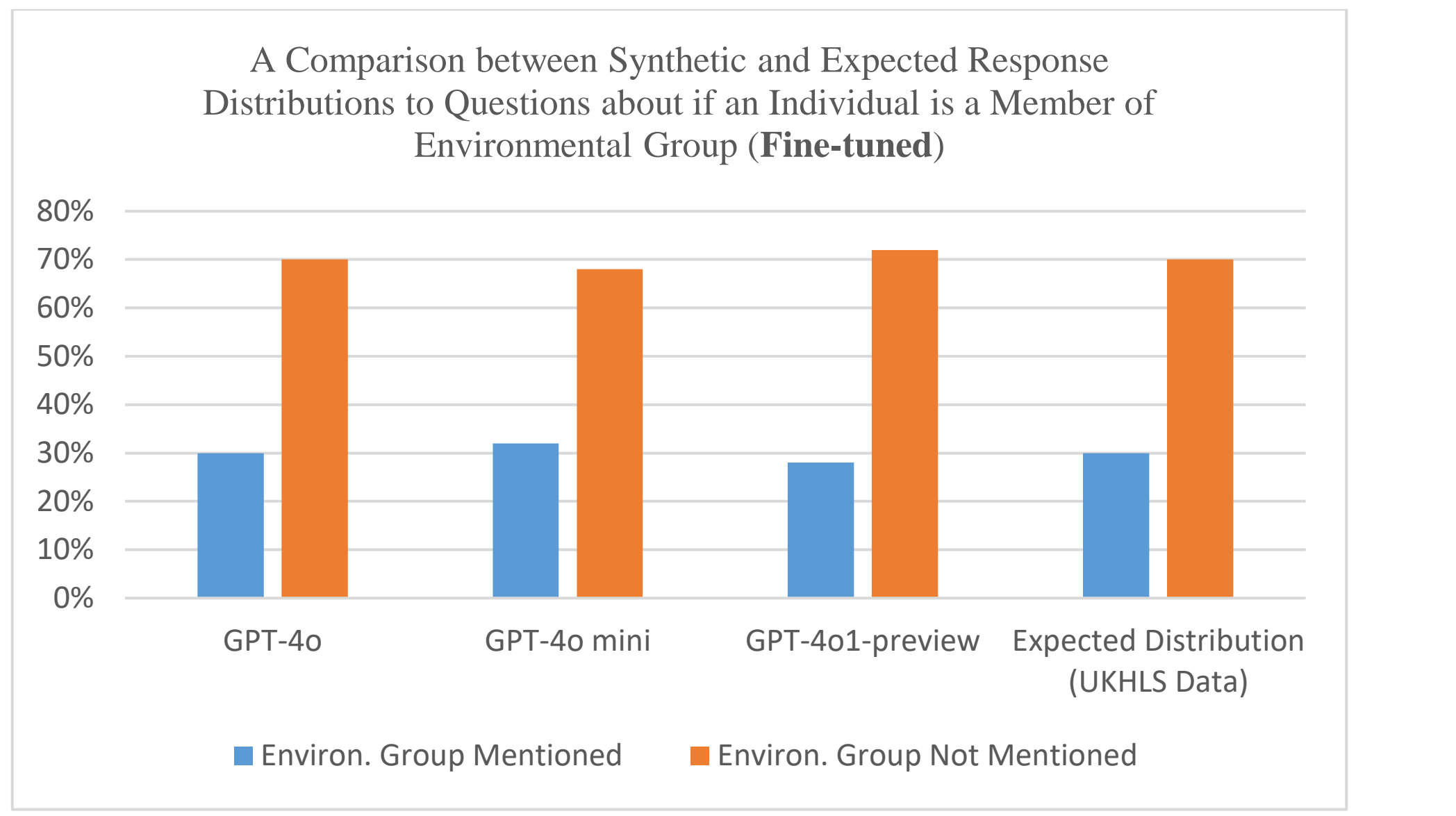

(b)

A Comparison between Synthetic and Expected Response Distributions to Questions about if an Individual is a Member of Environmental Group (**Pre-trained**)

80%
70%
60%
50%
40%
30%
20%
10%
0%

GPT-4o
GPT-4o mini
GPT-4o1-preview
Expected Distribution (UKHLS Data)

Environ. Group Mentioned
Environ. Group Not Mentioned

**Fig. A9** Graphs comparing the synthetic and expected response distributions to a question asking about if an individual is a member of an environmental organization.

(a)

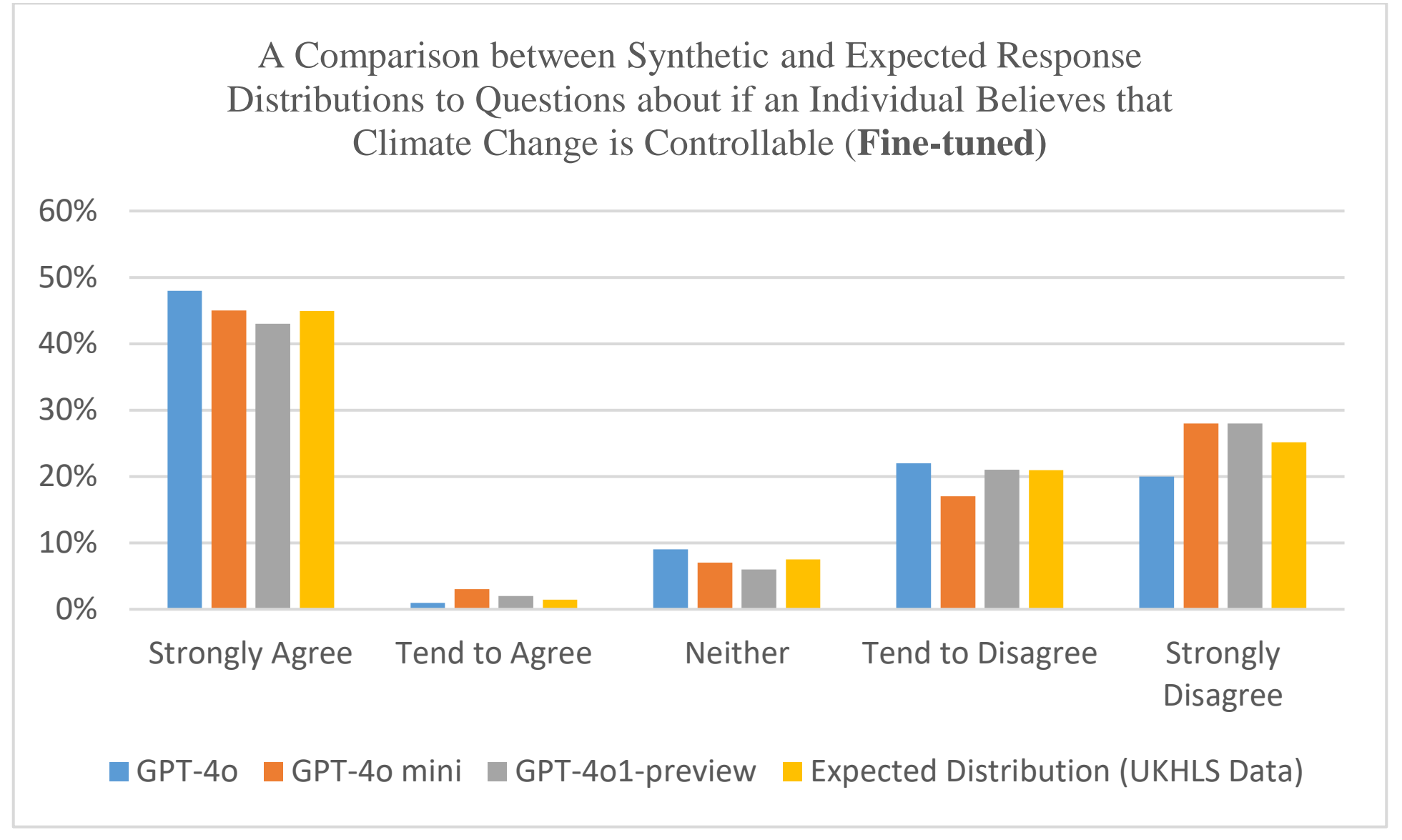

(b)

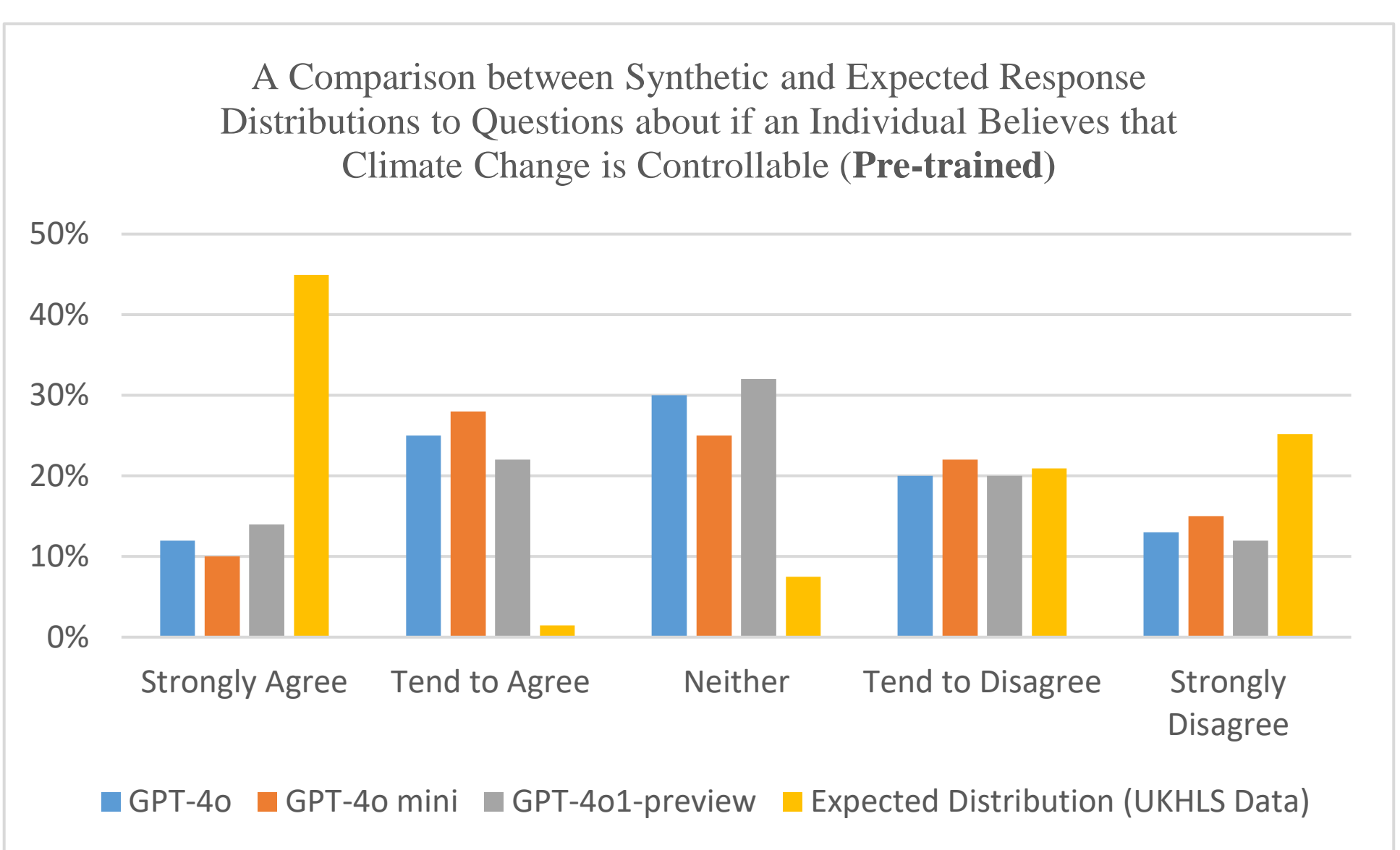

**Fig. A10** Graphs comparing the synthetic and expected response distributions to a question asking about if an individual believes that climate change is controllable: (a) fine-tuned (b) without fine-tuning.

## Appendix 2: Tables

**Table A1** An evaluation of the fine-tuned GPT-4o, GPT-4o-mini, and GPT-4o1-preview models using four key metrics: Chi-Square test scores, cosine similarity, Jaccard Index, and the Kullback-Leibler Divergence (KL Divergence). These models' synthetic distributions are compared against expected distributions sampled from the UKHLS datasets.

| Question | Model (**Fine-tuned**) | Chi-Square Test Scores | Cosine Similarity | Jaccard Index | KL-divergence |
|---|---|---|---|---|---|
| | GPT-4o | 0.85 | 0.96 | 0.72 | 0.12 |
| Describe your lifestyle | GPT-4o mini | 1.04 | 0.95 | 0.74 | 0.1 |
| | GPT-4o1-preview | 0.91 | 0.94 | 0.7 | 0.09 |
| | GPT-4o | 0.76 | 0.93 | 0.71 | 0.15 |
| Climate Change Impact | GPT-4o mini | 1.2 | 0.98 | 0.75 | 0.11 |
| | GPT-4o1-preview | 0.92 | 0.92 | 0.69 | 0.13 |
| | GPT-4o | 0.88 | 0.91 | 0.73 | 0.12 |
| Personal Impact on Climate | GPT-4o mini | 0.89 | 0.97 | 0.74 | 0.09 |
| | GPT-4o1-preview | 1.01 | 0.93 | 0.71 | 0.14 |
| | GPT-4o | 0.87 | 0.95 | 0.72 | 0.11 |
| Willing to Pay | GPT-4o mini | 0.85 | 0.92 | 0.71 | 0.09 |
| | GPT-4o1-preview | 0.91 | 0.94 | 0.72 | 0.11 |
| | GPT-4o | 1.04 | 0.91 | 0.73 | 0.14 |
| Personal Change | GPT-4o mini | 1.04 | 0.94 | 0.72 | 0.13 |
| | GPT-4o1-preview | 0.89 | 0.91 | 0.74 | 0.09 |
| | GPT-4o | 1.01 | 0.92 | 0.73 | 0.11 |
| Environ. Disaster | GPT-4o mini | 1.04 | 0.95 | 0.71 | 0.09 |
| | GPT-4o1-preview | 0.91 | 0.96 | 0.72 | 0.12 |
| | GPT-4o | 0.88 | 0.93 | 0.7 | 0.09 |
| Green Tariff | GPT-4o mini | 0.91 | 0.91 | 0.73 | 0.11 |
| | GPT-4o1-preview | 0.85 | 0.96 | 0.72 | 0.14 |
| | GPT-4o | 0.76 | 0.93 | 0.73 | 0.12 |
| Pollution | GPT-4o mini | 0.91 | 0.95 | 0.7 | 0.11 |
| | GPT-4o1-preview | 1.01 | 0.93 | 0.72 | 0.1 |
| | GPT-4o | 0.76 | 0.95 | 0.71 | 0.13 |
| Environ. Group | GPT-4o mini | 1.04 | 0.91 | 0.73 | 0.09 |
| | GPT-4o1-preview | 0.91 | 0.96 | 0.71 | 0.14 |
| | GPT-4o | 0.92 | 0.93 | 0.74 | 0.12 |
| Climate Change Control | GPT-4o mini | 0.91 | 0.95 | 0.73 | 0.11 |
| | GPT-4o1-preview | 0.92 | 0.91 | 0.72 | 0.09 |

**Table A2** An evaluation of the **pre-trained** GPT-4o, GPT-4o-mini, and GPT-4o1-preview models using four key metrics: Chi-Square test scores, cosine similarity, Jaccard Index, and the Kullback-Leibler Divergence (KL Divergence). These models' synthetic distributions are compared against expected real-world distributions sampled from the UKHLS datasets.

| Question | Model (**Pre-trained**) | Chi-Square Test Scores | Cosine Similarity | Jaccard Index | KL-divergence |
|---|---|---|---|---|---|
| | GPT-4o | 1.12 | 0.92 | 0.68 | 0.18 |
| Describe your lifestyle | GPT-4o mini | 1.25 | 0.91 | 0.69 | 0.16 |
| | GPT-4o1-preview | 1.1 | 0.9 | 0.7 | 0.14 |
| | GPT-4o | 1.15 | 0.89 | 0.65 | 0.19 |
| Climate Change Impact | GPT-4o mini | 1.33 | 0.91 | 0.68 | 0.15 |
| | GPT-4o1-preview | 1.2 | 0.88 | 0.69 | 0.17 |
| | GPT-4o | 1.05 | 0.87 | 0.66 | 0.13 |
| Personal Impact on Climate | GPT-4o mini | 1.18 | 0.92 | 0.67 | 0.18 |
| | GPT-4o1-preview | 1.3 | 0.9 | 0.68 | 0.15 |
| | GPT-4o | 1.17 | 0.91 | 0.7 | 0.16 |
| Willing to Pay | GPT-4o mini | 1.21 | 0.89 | 0.68 | 0.19 |
| | GPT-4o1-preview | 1.1 | 0.92 | 0.65 | 0.14 |
| | GPT-4o | 1.2 | 0.9 | 0.69 | 0.17 |
| Personal Change | GPT-4o mini | 1.22 | 0.91 | 0.68 | 0.18 |
| | GPT-4o1-preview | 1.35 | 0.89 | 0.67 | 0.16 |
| | GPT-4o | 1.18 | 0.92 | 0.68 | 0.19 |
| Environ. Disaster | GPT-4o mini | 1.1 | 0.91 | 0.69 | 0.14 |
| | GPT-4o1-preview | 1.12 | 0.93 | 0.7 | 0.16 |
| | GPT-4o | 1.25 | 0.91 | 0.66 | 0.15 |
| Green Tariff | GPT-4o mini | 1.15 | 0.9 | 0.67 | 0.13 |
| | GPT-4o1-preview | 1.18 | 0.89 | 0.65 | 0.18 |
| | GPT-4o | 1.2 | 0.87 | 0.68 | 0.19 |
| Pollution | GPT-4o mini | 1.11 | 0.88 | 0.69 | 0.14 |
| | GPT-4o1-preview | 1.16 | 0.9 | 0.68 | 0.15 |
| | GPT-4o | 1.12 | 0.91 | 0.7 | 0.17 |
| Environ. Group | GPT-4o mini | 1.25 | 0.92 | 0.68 | 0.19 |
| | GPT-4o1-preview | 1.18 | 0.93 | 0.69 | 0.16 |
| | GPT-4o | 1.2 | 0.9 | 0.67 | 0.14 |
| Climate Change Control | GPT-4o mini | 1.22 | 0.91 | 0.66 | 0.18 |
| | GPT-4o1-preview | 1.1 | 0.89 | 0.65 | 0.17 |